\documentclass{article}
\usepackage{iclr2027_conference,times}

\usepackage{amsmath,amsfonts,bm}

\def\eqref#1{equation~\ref{#1}}

\def\1{\bm{1}}

\DeclareMathAlphabet{\mathsfit}{\encodingdefault}{\sfdefault}{m}{sl}
\SetMathAlphabet{\mathsfit}{bold}{\encodingdefault}{\sfdefault}{bx}{n}

\usepackage[hidelinks]{hyperref}
\usepackage{url}
\usepackage{booktabs}
\usepackage{graphicx}
\usepackage{amssymb}
\usepackage{xcolor}
\usepackage{multirow}
\usepackage{enumitem}
\usepackage{fontawesome5}

\definecolor{linkblue}{RGB}{0,82,155}

\newcommand{\NumScenes}{300}
\newcommand{\GptTemplateDisp}{97}
\newcommand{\ClipLTemplateDisp}{48}
\newcommand{\BlipItmPitchDisp}{168}

\newcommand{\GptBestView}{1.20}
\newcommand{\BgIdentRate}{13}

\newcommand{\NRhoAlignPeakGtOne}{19}
\newcommand{\NRhoAlignCiGtOne}{19}
\newcommand{\NRhoAlignLogoGtOne}{19}
\newcommand{\NRhoAlignSampleGtOne}{19}
\newcommand{\NAlign}{19}

\newcommand{\MedSigPairs}{8}
\newcommand{\NPairsTotal}{15}
\newcommand{\GptTmplSE}{88}
\newcommand{\GptTmplOther}{117}
\newcommand{\GptScrSE}{56}
\newcommand{\GptScrOther}{56}
\newcommand{\ClipLTemplateReal}{27}
\newcommand{\GptTemplateReal}{61}

\newcommand{\OpenClipLScr}{58}
\newcommand{\BigGScr}{66}
\newcommand{\OpenClipLTmplRho}{6.53}
\newcommand{\BigGTmplRho}{3.08}

\newcommand{\TmplNounDisp}{10}
\newcommand{\TmplFrameDisp}{13}

\newcommand{\NounSweepN}{19}
\newcommand{\BestTmplInfl}{17}
\newcommand{\BestTmplInflMax}{39}
\newcommand{\DefaultTmplRank}{200}
\newcommand{\NumTemplates}{337}

\newcommand{\EnvEvCorr}{0.84}
\newcommand{\EnvEvSpan}{5.25}

\newcommand{\AuditPointEstimateReversals}{18}
\newcommand{\AuditPointwiseReversals}{9}
\newcommand{\AuditSimultaneousReversals}{0}
\newcommand{\AuditNBootstrap}{2000}
\newcommand{\AuditNContrasts}{110,295}
\newcommand{\AuditMaxCritical}{4.365}
\newcommand{\AuditGptDrops}{127}
\newcommand{\AuditGptTies}{84}
\newcommand{\AuditGptRises}{89}
\newcommand{\AuditGptZeroPairs}{11}
\newcommand{\AuditGptNonzeroPairs}{289}
\newcommand{\AuditGptSensitivityLo}{53.7}
\newcommand{\AuditGptSensitivityHi}{57.3}
\newcommand{\AuditGptDropPct}{42.3}
\newcommand{\AuditGptDropLo}{36.0}
\newcommand{\AuditGptDropHi}{48.3}
\newcommand{\AuditGptTiePct}{28.0}
\newcommand{\AuditGptTieLo}{23.0}
\newcommand{\AuditGptTieHi}{33.3}
\newcommand{\AuditGptRisePct}{29.7}
\newcommand{\AuditGptRiseLo}{24.7}
\newcommand{\AuditGptRiseHi}{34.7}
\newcommand{\AuditGptCreditPct}{56.3}
\newcommand{\AuditGptCreditLo}{51.0}
\newcommand{\AuditGptCreditHi}{61.3}

\title{Frozen Scenes, Shifting Winners: Configuration Fragility in Text-to-3D Evaluation}

\author{Anson Y. Lam\textsuperscript{$\ast$}, Shuqing Li\textsuperscript{$\ast,\dagger$} \& Michael R. Lyu \\
Department of Computer Science and Engineering \\
The Chinese University of Hong Kong \\
Hong Kong, China \\
\{yflam1,sqli21,lyu\}@cse.cuhk.edu.hk \\[-2em]
}

\iclrfinalcopy %
\begin{document}

\maketitle

\lhead{}
{\renewcommand{\thefootnote}{\fnsymbol{footnote}}%
    \setcounter{footnote}{1}\footnotetext{Equal contribution.}%
    \setcounter{footnote}{2}\footnotetext{Corresponding author.}%
    \setcounter{footnote}{0}}

\begin{center}
    \textcolor{linkblue}{\faHome}~\href{https://yflam1.github.io/frozen-scenes-shifting-winners/}{\textcolor{linkblue}{\textbf{Project Page}}}
    \quad
    \textcolor{linkblue}{\faGithub}~\href{https://github.com/yflam1/frozen-scenes-shifting-winners}{\textcolor{linkblue}{\textbf{Code}}}
\end{center}

\begin{abstract}
    Can a text-to-3D leaderboard change when every generated scene stays fixed?
    We audit this question for rendered-image evaluation, where camera settings and caption wording become part of the measurement protocol.
    Across \NumScenes{} frozen scenes from six generators, we vary eight render and caption factors for 19 alignment evaluators plus one perceptual-quality control, then test four targeted scene degradations.
    Peak configuration variance exceeds between-generator variance for \NRhoAlignPeakGtOne/\NAlign{} alignment evaluators, with prompt-bootstrap lower bounds above $1$ for \NRhoAlignCiGtOne/\NAlign{}.
    Rankings are more stable than scores, yet \AuditPointEstimateReversals/\NAlign{} evaluators change their point-estimate winner under some configuration.
    Pairwise protocol margin envelopes show which comparisons keep their direction across the tested settings.
    Selected pairs have opposite pointwise intervals, but no reversal survives simultaneous inference over the full search.
    Thus the observed winner changes are descriptive, not confirmed changes in generator superiority.
    Sensitivity remains separate: no evaluator exceeds $71\%$ tie-adjusted directional discrimination on layout scrambling, and a preregistered blinded human study classifies part of the configuration variance as metric artifact.
    The audit separates score stability, decision uncertainty, and targeted sensitivity, and recommends reporting \emph{(generator, score, card ID)} with protocol-dependent comparisons and selection-aware uncertainty.
\end{abstract}

\section{Introduction}
\label{sec:intro}

A text-to-3D scene paper can change its reported winner without changing any generated scene.
A common evaluation recipe renders one view, wraps the prompt in a caption template, and asks a vision-language model whether they match.
Camera, lighting, and wrapper wording then become part of the evaluator.
The scientific question is which scene-generator comparisons remain credible when those choices are made explicit.

\begin{figure}[ht]
    \centering
    \includegraphics[width=0.80\linewidth]{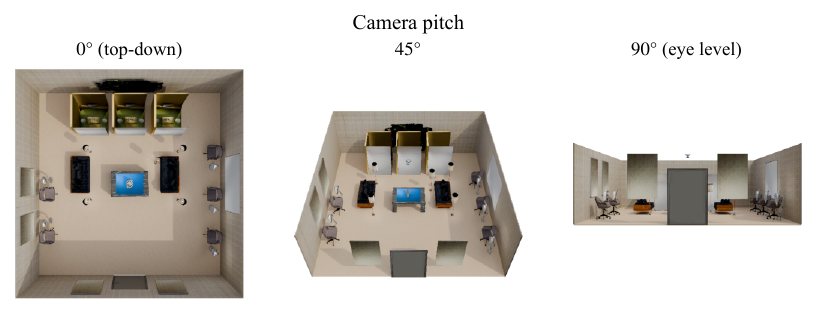}
    \caption{\textbf{The artifact stays fixed, but the measurement changes.}
        Three camera pitches reveal different evidence from one frozen room.
        Our audit holds all 300 scene artifacts fixed and varies the rendering or caption protocol.
        Winner changes are measured across all 50 prompts, not inferred from this pictured room, and are separated from statistical confirmation in Section~\ref{sec:results-rank}.}
    \label{fig:thesis}
\end{figure}

This matters because generated indoor scenes are becoming useful, navigable artifacts for design, media, and embodied AI~\citep{holodeck,instructscene,reason3d}.
As generators improve, a metric that responds to an incidental view or wrapper can steer comparisons toward the wrong model.
Building on SceneCritic's view-dependent VLM-judge findings~\citep{scenecritic} and 3D-DefectBench's pipeline audit for human-referenced defect detection~\citep{defectbench}, we ask a narrower decision question: within a named evaluator, how much do plausible render and caption settings move scores and generator comparisons on the same fixed scenes?

We audit this question under a fixed-artifact counterfactual design.
Holding all generated scenes fixed, we vary one measurement-protocol factor at a time across 19 alignment evaluators and one perceptual-quality control.
Following measurement theory~\citep{measuredatasetdiversity,whatmakesgoodmetric}, this design separates reliability under configuration choices from sensitivity to targeted scene changes.
We call a change in the unique top-ranked generator a \emph{same-scene point-estimate winner reversal} when only the evaluator configuration changes.
The term describes the reported decision, not statistically established generator superiority.

The audit gives three main findings.
First, peak configuration variance exceeds between-generator variance for \NRhoAlignPeakGtOne{} of \NAlign{} alignment evaluators, with bootstrap lower bounds above $1$ for \NRhoAlignCiGtOne{}.
Second, rankings are more stable than scores, but \AuditPointEstimateReversals{} evaluators change their point-estimate winner.
Selected generator pairs even yield opposite pointwise intervals, although no reversal survives the full-search simultaneous band.
Third, rank stability does not imply scene understanding: no evaluator exceeds $71\%$ tie-adjusted discrimination on layout scrambling, and a preregistered blinded human study classifies part of the camera-driven score movement as metric artifact (Appendix~\ref{app:human-study}).
Pairwise protocol margin envelopes connect these observations by identifying which sampled comparisons keep their direction across the declared configuration set.

Our contributions are:
\begin{itemize}[leftmargin=1.4em,itemsep=1pt,topsep=2pt]
    \item \textbf{A fixed-artifact audit design} covering \NumScenes{} frozen scenes, six generators, 19 alignment evaluators plus one control, eight render and caption factors, and four targeted degradations.
    \item \textbf{A decision analysis} that separates score drift, descriptive winner changes, pairwise protocol margins, pointwise uncertainty, and full-search simultaneous inference.
    \item \textbf{A reporting contract} centered on a versioned Evaluator Configuration Card, so each reported score can be tied to the render and caption choices behind it.
\end{itemize}

The practical outcome is an audit before any superiority claim, and Section~\ref{sec:discussion} turns it into a reporting checklist.

\section{Related Work}
\label{sec:related}

\paragraph{Rendered-image evaluation for text-to-3D scenes.}
Text-to-scene generators span procedural construction, diffusion layouts, VLM optimization, and reasoning pipelines~\citep{holodeck,diffuscene,layoutvlm,reason3d}.
Their alignment evaluation often reduces a 3D scene to one rendered image and one text query: Holodeck uses top-down CLIPScore~\citep{holodeck}, Holodeck~2.0 scores perspective renders~\citep{holodeck2}, and HSM adds BLIP-2, Long-CLIP, and VQAScore~\citep{hsm}.
Because rendering settings and caption wrappers are often under-specified, the metric name alone does not define the measurement.

\paragraph{Metric audits and measurement evidence.}
CLIPScore made reference-free image-text scoring common~\citep{clipscore}, but CLIP-like encoders struggle with relations, counting, and viewpoint changes~\citep{aro,clipcount10,tdbench}.
Newer metrics use question answering or VLM scoring~\citep{tifa,vqascore}, while audits reveal score artifacts, prompt sensitivity, and controlled-degradation blind spots~\citep{metricmirage,whatmakesgoodmetric,diagnosticaudit}.
Evaluator-side unreliability is documented as well: visual biases sway LVLM judges~\citep{foolinglvlm}, evaluator VLMs have blind spots~\citep{seeingbelieving}, and rankings prove more stable than judge scores~\citep{vlmrankscore}.
One response builds more human-aligned judges~\citep{genarena}, while our audit quantifies evaluator unreliability against the generator decision signal on fixed artifacts, with search-corrected simultaneous inference.
AutoMetrics is closest in spirit because it separates stability under same-quality perturbations from sensitivity to worse-quality perturbations~\citep{autometrics}.
RLiable supplies uncertainty-aware comparison tools~\citep{rliable}, and Evaluation Cards records evaluation-run provenance~\citep{evalcards}.
Our card records render and caption choices for published vision evaluators on frozen 3D scene artifacts.

\paragraph{Fine-grained 3D evaluation and closest boundaries.}
Object and mesh benchmarks use multi-view aggregation or learned evaluators~\citep{t3bench,threedgenbench,db3dme}, while scene benchmarks improve spatial and trajectory-based evidence~\citep{sceneeval,worldscore}.
SceneCritic is the closest same-artifact instability study: it reports repeated-call and view-dependent score variance and method-ranking reversals for one Gemini-2.5-Pro judge, then proposes a human-validated symbolic floor-plan evaluator~\citep{scenecritic}.
3D-DefectBench studies the full VLM-based defect-detection pipeline, crossing judge, camera protocol, visual input, and prompt schema against human fine-grained defect labels~\citep{defectbench}.
Our contribution is different in scope.
We vary render and caption settings within each audited evaluator and trace score movement to generator decisions.
The audit spans \NAlign{} alignment evaluators, eight factors, and 387 shared protocols on identical artifacts, finds peak configuration variance exceeding between-generator variance for \NRhoAlignPeakGtOne{} of \NAlign{} evaluators as a point estimate, yields zero confirmed reversals in decision-level protocol margins under selection-aware simultaneous inference over \AuditNContrasts{} searched contrasts, and adds four targeted scene degradations as necessary-condition probes.
Thus the paper supports a reporting contract for score comparability rather than a new scene-quality metric.
Appendix~\ref{app:closest-work} gives the nearest-work boundary and broader references, and Appendix~\ref{app:card-scope} defines the card fields.

\section{Experimental Setup}
\label{sec:method}

\subsection{Audit design}
\label{sec:method-overview}

We ask whether common evaluator choices change conclusions when the generated scene itself is frozen.
For prompt $p$ and generator $m$, the generated structured scene is $s=m(p)$.
A renderer produces an image $x=\mathcal{R}(s,c)$, an evaluation caption $t=\tau(p')$ wraps either the original prompt or a paraphrase, and evaluator $e$ returns $\sigma_e(x,t)$.
We call the full measurement protocol $\pi=(\mathcal{R},c,\tau,p',e,\text{preprocessing},\text{score scale})$ and write $q_e(m,p;\pi)$ for the score of the fixed scene $m(p)$.
This notation makes the audited object explicit: the scene is fixed, while the protocol that turns that scene into a number is varied.

The audit has two parts.
A \textbf{configuration check} changes one evaluator-side choice at a time while holding the scene and all other choices at the default $c_0$, following the reliability side of measurement-theoretic audits~\citep{measuredatasetdiversity,whatmakesgoodmetric,t2iscorescore}.
A \textbf{directional degradation check} edits the structured scene in a targeted way and asks whether scores drop on average, giving intervention-specific sensitivity evidence.
The first part measures protocol sensitivity, while the second tests response to specific interventions rather than general validity.
Neither is sufficient alone, since a stable score can track the wrong signal and response to one degradation need not establish broad validity.
The one-factor-at-a-time design keeps the render count linear and estimates each factor conditional on the default settings.
A preregistered two-by-two caption-by-pitch factorial on the same \NumScenes{} scenes now estimates the largest candidate interaction, and its size is evaluator-dependent (Appendix~\ref{app:followup-factorial}).

\subsection{Scenes, protocols, and evaluators}
\label{sec:method-data}

We use $N=50$ prompts drawn with a fixed stratified seed from SceneEval, InstructScene, Holodeck, and Scenethesis~\citep{sceneeval,instructscene,holodeck,constraintir}.
The sample weights SceneEval more heavily because its object annotations match the content probes, and the resulting intervals are diagnostic prompt-bootstrap summaries rather than population estimates.
Six generators produce the frozen scenes: Scenethesis, Holodeck, HSM, LayoutVLM, I-Design, and Reason-3D~\citep{constraintir,holodeck,hsm,layoutvlm,idesign,reason3d}.
These systems span constraint-IR synthesis, LLM procedural construction, hierarchical motifs, differentiable VLM optimization, agentic design, and reasoning-model planning.
All emit structured object-level scenes with asset meshes, poses, and boxes, giving \NumScenes{} fixed scenes for the audit and enabling object-level perturbations.
Methods that accept a configurable text backbone use the recorded identifier \texttt{gemini-3-flash-preview} at temperature $0.7$.
Each generated scene is saved once and reused throughout the audit.

Every scene is rendered through one Blender Cycles pipeline.
Transparent-film rendering lets us composite backgrounds after lighting, which decouples the background factor from illumination.
Auto-framing keeps the scene box fixed in the image as focal length changes, and the dollhouse wall convention keeps indoor content visible under oblique views.
We sweep eight evaluator-configuration factors independently: resolution, focal length, lighting, background, pitch, yaw, caption template, and paraphrase (Table~\ref{tab:factors}).
The default $c_0$ is a $512$ px render, $50$ mm focal length, city environment map, white background, top-down pitch, default yaw, the default caption template, and the original prompt.
The shared sweep has 9 resolution levels, 7 focal lengths, 8 lighting maps, 10 backgrounds, 7 pitch values, 8 yaw values, and 337 caption templates.
Paraphrases are scored only at the default render because their levels are prompt-specific strings.
Yaw is swept at a fixed $45^\circ$ pitch because yaw at the default top-down pitch is mostly an in-plane rotation.
The camera sweep also defines a best-view test by taking the largest score over the tested poses.
These factors are protocol choices, not all strict nuisance transformations, because captions can alter textual assertions and camera or resolution settings can change visible evidence.
Table~\ref{tab:factors} gives the full level inventory, so the main text can focus on what each family of choices estimates.

\label{sec:method-text}
Caption templates are all source-to-sink paths through a small grammar, giving 337 wrappers including the default \emph{``a top-down view of \{prompt\}''} and the raw prompt (Appendix~\ref{app:templates}).
Paraphrases use the frozen file of 10 intended meaning-preserving rewrites per prompt.
Duplicate caption strings collapse in scoring, giving 8--11 distinct captions per prompt.
Two hosted auditors accept only about half of the 500 rewrites as meaning preserving, so paraphrase dispersion is treated as a stress test of the full text pipeline rather than a primary invariance claim (Appendix~\ref{app:paraphrase-audit}).

We audit 19 text-scene alignment evaluators plus one scoped perceptual-quality control (Table~\ref{tab:evaluators}).
The set covers CLIP-family encoders used in text-to-3D scene papers, SigLIP, BLIP-2, BLIP-VQA, VQAScore, and a hosted judge.
The CLIP-family sweep varies architecture, training data, patch size, and native resolution, which helps separate a single-checkpoint failure from a broader pattern in metrics used by the field.
The hosted-judge row uses \texttt{gpt-4o-2024-11-20}.
The scoring client records request and parsing exceptions as zero scores.
Every stored zero cell has since been re-scored, and a three-policy sensitivity analysis over stored, dropped, and re-scored zero cells appears in Appendix~\ref{app:followup-zeros}.
Appendix~\ref{app:hosted-provenance} bounds the effect of stored zero scores with a re-scoring analysis and a deterministic sensitivity range.
CLIP-IQA is a prompt-free image-quality control and is excluded from alignment-only headline counts.

\subsection{Content probes}
\label{sec:method-content}

The content probes construct a paired stress-test scene $s^-$ and render it at $c_0$ against the original prompt.
Across prompts that specify the targeted property, a responsive evaluator should tend to score the original at least as high as the edited scene.
Because some generated originals may already be wrong and some prompts are generic, this is a dataset-level diagnostic expectation rather than pairwise human ground truth.
Directional discrimination credits a strict drop as 1, an exact tie as 0.5, and a rise as 0, so chance is $50\%$ under symmetric random ordering.

The four single-shot probes are \emph{keep-half}, \emph{biggest-only}, \emph{scrambled}, and \emph{worst-objects}.
\emph{Keep-half} removes a random half of objects, and \emph{biggest-only} keeps only the largest object.
\emph{Scrambled} relocates and reorients every object while preserving the object set, making it the cleanest layout probe.
\emph{Worst-objects} replaces each asset with the lowest-ranked retrieval candidate under the generator's own asset score while preserving label, pose, size, and count.
Worst-objects is least clean because CLIP-family evaluators may partly regain the retrieval signal used to select the bad asset.
For that reason, the results emphasize scrambling as the necessary-condition probe for spatial arrangement sensitivity.

\subsection{Statistics and uncertainty}
\label{sec:method-analysis}

For a factor $f$ with levels $\ell$, let $\sigma_{s,\ell}$ be the score of scene $s$ at level $\ell$, $\bar\sigma_s$ its mean over levels, and $\mathbb{E}_s$ the scene average.
\textbf{Dispersion} measures score movement as $D^{(f)}=\mathbb{E}_{s}\!\left[(\max_\ell\sigma_{s,\ell}-\min_\ell\sigma_{s,\ell})/\bar\sigma_s\right]$.
\textbf{Configuration-to-signal ratio} compares within-scene configuration variance with between-generator mean-score variance at the default, $\rho^{(f)}=\mathbb{E}_{s}\!\left[\operatorname{Var}_\ell\sigma_{s,\ell}\right]/\operatorname{Var}_{m}\!\left(\mathbb{E}_{p}\sigma_{m,p,c_0}\right)$.
The denominator is the variance over the six generator means after averaging prompts at $c_0$.
Thus $\rho>1$ means configuration levels contribute more variance than generator means on that evaluator's own scale.
It is not a statement that a configuration is more important than generator quality in a human sense.
We report template $\rho$ and peak $\rho$ per evaluator.
\textbf{Rank stability} uses Kendall's coefficient of concordance $W$ over factor levels, where $W=1$ means the same order at every level.
A same-scene point-estimate winner change occurs when two levels of a factor have different unique top generators while every $m(p)$ is fixed.
\textbf{Best-view inflation} is the median over scenes of the largest tested-view score divided by the default-view score, restricted to positive baselines.
Together these statistics separate score stability, decision robustness, and targeted probe sensitivity, which a single leaderboard number would collapse.

Generator comparisons use the same fixed scenes.
For evaluator $e$, let $\Pi_e$ contain the baseline and all shared-level protocols, excluding scene-specific paraphrases.
For generators $m,n$, define $L_e(m;\pi)=N^{-1}\sum_p q_e(m,p;\pi)$ and $\Delta_{e,mn}(\pi)=L_e(m;\pi)-L_e(n;\pi)$.
If $u_{e,m}(\pi)=L_e(m;\pi)-L_e(m;\pi_0)$, then
\begin{equation}
    \Delta_{e,mn}(\pi)=\Delta_{e,mn}(\pi_0)+u_{e,m}(\pi)-u_{e,n}(\pi).
    \label{eq:margin-shift}
\end{equation}
Common protocol shifts cancel, so high dispersion or high $\rho$ is not by itself a changed comparison.
We summarize the finite audit set by the protocol margin envelope
\begin{equation}
    I_{e,mn}(\Pi_e)=\left[\min_{\pi\in\Pi_e}\Delta_{e,mn}(\pi),\;\max_{\pi\in\Pi_e}\Delta_{e,mn}(\pi)\right].
    \label{eq:margin-envelope}
\end{equation}
A strictly positive or negative envelope preserves the pair ordering across tested protocols, while a sign crossing records configuration dependence.
This envelope is a descriptive range, not a confidence interval, and it makes no claim about untested factor combinations, new prompt samples, or human scene quality.

Prompt bootstrap gives pointwise descriptive intervals for dispersion, $\rho$, directional discrimination, and selected pairwise margins.
The bootstrap resamples the 50 prompt blocks and keeps the six generators, evaluators, and configurations jointly aligned within each prompt.
For winner reversals, we also use an approximate simultaneous bootstrap band over the full searched family of 19 alignment evaluators, 15 generator pairs, and 387 shared configurations (Appendix~\ref{app:selection-inference}).
This family contains \AuditNContrasts{} contrasts and excludes the image-only control and prompt-specific paraphrase levels.
Exact aggregate ties remain ties.
A point-estimate top-1 change alone carries no significance claim, and even opposite pairwise intervals do not prove either generator beats all five alternatives.
Paraphrases are excluded from the rank-stability and winner-reversal search because their levels are prompt-specific strings rather than shared protocols.
Appendix~\ref{app:followup-zeros} reports a reanalysis restricted to the rewrites that both auditors jointly accept.

\section{Results}
\label{sec:results}

\begin{figure}[t]
    \centering
    \includegraphics[width=0.87\linewidth]{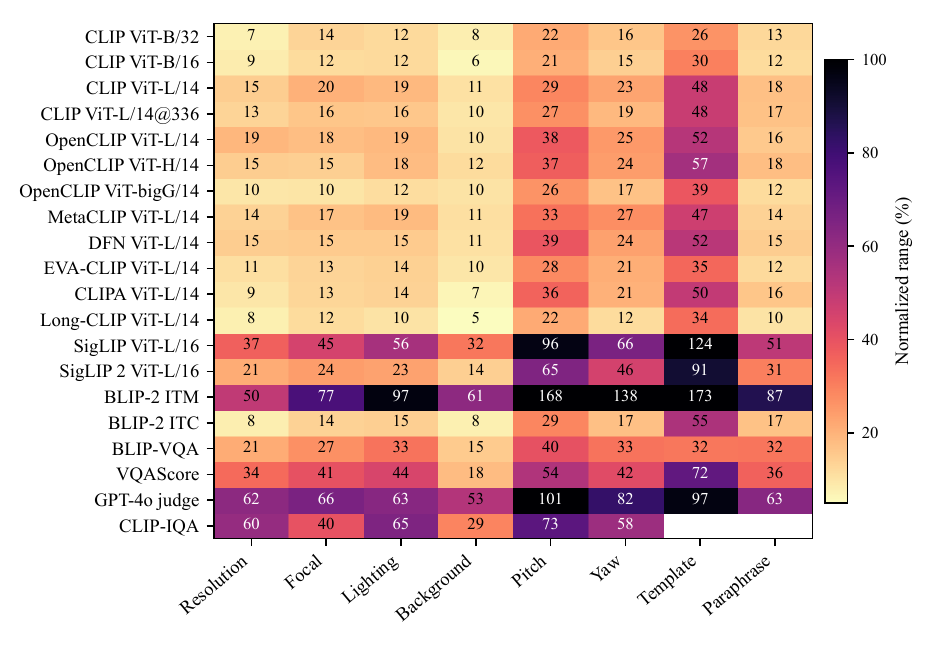}
    \vspace{-1.5em}
    \caption{Score dispersion (normalized range, \%) across evaluators and configuration factors in default-conditioned one-factor-at-a-time sweeps.
        Cells show the true values, and colors are clipped at $100\%$.
        Caption template is the largest factor for 17 of 19 alignment evaluators, while pitch and lighting dominate the render factors.
        The prompt-free CLIP-IQA control has no caption measurements.}
    \label{fig:heatmap}
    \vspace{-1.5em}
\end{figure}

The scenes stay fixed throughout the audit, yet scores and point-estimate winners depend on how those scenes are presented to an evaluator.
Table~\ref{tab:decision-summary} separates score movement, generator comparisons, and targeted content sensitivity.
Figure~\ref{fig:heatmap} shows the factor-by-evaluator dispersion behind the score-movement result.
Table~\ref{tab:headline} reports the per-evaluator decision measures, including the render-only quality control.
Appendix~\ref{app:per-evaluator} gives the complete per-factor results.

\begin{table}[t]
    \caption{Decision-level results from 50 prompts, six generators, and \NumScenes{} fixed scenes, scored by \NAlign{} alignment evaluators and one render-only control.
        Configuration sweeps change one factor at a time from a default, so these results do not measure factor interactions.}
    \label{tab:decision-summary}
    \centering
    \small
    \newcommand{\auditquestion}[1]{\parbox[t]{0.28\linewidth}{\raggedright #1}}
    \newcommand{\auditresult}[1]{\parbox[t]{0.11\linewidth}{#1}}
    \newcommand{\auditscope}[1]{\parbox[t]{0.51\linewidth}{\raggedright #1}}
    \begin{tabular}{@{}lll@{}}
        \toprule
        Audit question                                                           & Result                                               & Scope and interpretation                                                                         \\
        \midrule
        \auditquestion{Does configuration variance exceed generator separation?} & \auditresult{\NRhoAlignPeakGtOne/\NAlign{}}          & \auditscope{Peak $\rho>1$ for the six-generator field.}                                          \\ \addlinespace[2pt]
        \auditquestion{Does this persist under prompt resampling?}               & \auditresult{\NRhoAlignCiGtOne/\NAlign{}}            & \auditscope{Bootstrap lower bound for peak $\rho$ exceeds 1.}                                    \\ \addlinespace[2pt]
        \auditquestion{Does the point-estimate winner change?}                   & \auditresult{\AuditPointEstimateReversals/\NAlign{}} & \auditscope{Across tested factor levels, without claiming a confirmed reversal.}                 \\ \addlinespace[2pt]
        \auditquestion{Is an opposite pairwise ordering confirmed after search?} & \auditresult{\AuditSimultaneousReversals/\NAlign{}}  & \auditscope{Evaluators with a selected reversal under a band over \AuditNContrasts{} contrasts.} \\ \addlinespace[2pt]
        \auditquestion{Which pairwise orders survive all tested protocols?}      & \auditresult{0--7/15}                                & \auditscope{Strict orders per alignment evaluator over a finite set of 387 protocols.}           \\ \addlinespace[2pt]
        \auditquestion{Does scrambling lower the score?}                         & \auditresult{$\leq 71\%$}                            & \auditscope{Tie-adjusted ceiling across evaluators, including the render-only control.}          \\
        \bottomrule
    \end{tabular}
\end{table}

\begin{table}[t]
    \caption{Per-evaluator audit across four distinct questions.
        Template and peak $\rho$ compare configuration variance with between-generator variance in the six-generator field.
        Rank stability gives the minimum Kendall's $W$ and the maximum number of distinct point-estimate winners.
        Content accuracy averages four directional probes and is not a human-validity score, and best view is the median score inflation from the highest-scoring tested camera.
        $^\dagger$ marks a prompt-bootstrap lower bound above one for peak $\rho$ in all 19 alignment rows and the CLIP-IQA control (Appendix~\ref{app:per-evaluator}).}
    \label{tab:headline}
    \centering
    \begin{tabular}{@{}l cc cc cc@{}}
        \toprule
                           & \multicolumn{2}{c}{$\bm\rho$} & \multicolumn{2}{c}{\textbf{Rank stability}} & \multirow{3}{*}{\textbf{\begin{tabular}[c]{@{}c@{}}Content\\acc. (\%)\end{tabular}}} & \multirow{3}{*}{\textbf{\begin{tabular}[c]{@{}c@{}}Best\\view\end{tabular}}}      \\
        \cmidrule(lr){2-3}
        \cmidrule(lr){4-5}
        \textbf{Evaluator} & template                      & peak                                        & min $W$                                                                              & \#top-1                                                                      &  & \\
        \midrule
        CLIP ViT-B/32 & 6.80 & 13.89$^\dagger$ & 0.55 & 2 & 68 & 1.18\,$\times$ \\
CLIP ViT-B/16 & 9.35 & 13.01$^\dagger$ & 0.34 & 3 & 70 & 1.15\,$\times$ \\
CLIP ViT-L/14 & 7.04 & 7.08$^\dagger$ & 0.50 & 2 & 74 & 1.12\,$\times$ \\
CLIP ViT-L/14@336 & 4.95 & 4.95$^\dagger$ & 0.55 & 3 & 76 & 1.10\,$\times$ \\
OpenCLIP ViT-L/14 & 6.53 & 10.13$^\dagger$ & 0.52 & 2 & 73 & 1.08\,$\times$ \\
OpenCLIP ViT-H/14 & 12.96 & 12.99$^\dagger$ & 0.55 & 3 & 78 & 1.03\,$\times$ \\
OpenCLIP ViT-bigG/14 & 3.08 & 4.26$^\dagger$ & 0.50 & 3 & 75 & 1.04\,$\times$ \\
MetaCLIP ViT-L/14 & 3.48 & 4.22$^\dagger$ & 0.55 & 2 & 71 & 1.09\,$\times$ \\
DFN ViT-L/14 & 3.24 & 4.62$^\dagger$ & 0.80 & 2 & 78 & 1.05\,$\times$ \\
EVA-CLIP ViT-L/14 & 3.11 & 4.37$^\dagger$ & 0.62 & 3 & 71 & 1.05\,$\times$ \\
CLIPA ViT-L/14 & 3.51 & 4.29$^\dagger$ & 0.72 & 3 & 68 & 1.03\,$\times$ \\
Long-CLIP ViT-L/14 & 12.43 & 14.24$^\dagger$ & 0.67 & 2 & 79 & 1.05\,$\times$ \\
SigLIP ViT-L/16 & 6.89 & 7.94$^\dagger$ & 0.61 & 2 & 76 & 1.08\,$\times$ \\
SigLIP 2 ViT-L/16 & 6.55 & 6.64$^\dagger$ & 0.61 & 2 & 75 & 1.05\,$\times$ \\
BLIP-2 ITM & 19.42 & 19.42$^\dagger$ & 0.66 & 3 & 77 & 1.02\,$\times$ \\
BLIP-2 ITC & 12.11 & 12.11$^\dagger$ & 0.69 & 2 & 73 & 1.04\,$\times$ \\
BLIP-VQA & 5.26 & 13.90$^\dagger$ & 0.65 & 3 & 72 & 1.20\,$\times$ \\
VQAScore & 1.64 & 3.88$^\dagger$ & 0.51 & 2 & 74 & 1.29\,$\times$ \\
GPT-4o judge & 0.87 & 3.48$^\dagger$ & 0.77 & 1 & 77 & 1.20\,$\times$ \\
CLIP-IQA & -- & 3.49$^\dagger$ & 0.24 & 4 & 72 & 1.22\,$\times$ \\
                                                                                                                                                                                                                                                     \\[-1.2em]
        \bottomrule
    \end{tabular}
    \vspace{-1.5em}
\end{table}

{\small
\noindent\textbf{What this audit establishes.}
\textbf{Established:} the caption template is the largest tested factor for 17 of \NAlign{} alignment evaluators, and the hosted judge's template dispersion stays large under every zero-handling policy ($60.6\%$ as stored, $46.5\%$ with zero cells dropped, $47.6\%$ with zeros replaced, across common caption forms, Appendix~\ref{app:followup-zeros}).
Peak configuration variance exceeds between-generator variance for \NRhoAlignPeakGtOne{} of \NAlign{} alignment evaluators, with a prompt-bootstrap lower bound above one for \NRhoAlignCiGtOne.
Exactly one strict ordering survives the same full-search simultaneous band under which no reversal is confirmed, a loss for Holodeck (Scenethesis over Holodeck, for one evaluator).
A preregistered replication on 50 new disjoint prompts confirms the headline: 19 of 19 alignment evaluators change their point-estimate winner under both definitions (Appendix~\ref{app:n100}).
A preregistered $2\times2$ factorial estimates the caption-by-pitch interaction at $-0.523$ $[-0.670,-0.383]$ on the pinned evaluator (Appendix~\ref{app:followup-factorial}).
The generator-by-level component implicated in winner flips stays small, at most $5.33\%$ of total variance at the median within any swept factor, and every factor other than pitch stays at or below $0.98\%$.

\textbf{Not established (and what would establish it):} a confirmed winner reversal, since \AuditSimultaneousReversals{} of \NAlign{} evaluators survive the full-search simultaneous band, and only the preregistered larger sample can settle the four pinned contrasts.
Whether score shifts track what a human sees beyond the audited pairs. The completed blinded study gives a first artifact verdict (Appendix~\ref{app:human-study}).
Configuration effects beyond the swept indoor Blender pipeline, which wider sweeps would have to show.
Planning estimates put that confirmation at roughly 75 to 100 prompts per contrast at Holm-corrected $\alpha$, which the paper commits to on an independent sample (Appendix~\ref{app:followup-power}).}

\subsection{Render-side sensitivity: the camera moves the score}
\label{sec:results-render}

Among render factors, camera pitch and lighting produce the largest default-conditioned shifts (Figure~\ref{fig:heatmap}).
Changing pitch from top-down to eye level moves BLIP-2 ITM by \BlipItmPitchDisp\% and the CLIP encoders by $21$--$39\%$, consistent with their documented top-down blind spot~\citep{tdbench}.
Across eight lighting maps spanning about \EnvEvSpan{} exposure stops, brightness correlates $r=\EnvEvCorr$ with the mean score shift (Table~\ref{tab:envmaps}).
This association does not isolate brightness from other map properties.
Resolution and focal length move CLIP-family scores by roughly $7$--$20\%$ each, while background is mildest at about $10\%$.
The background result is partly mechanical, since compositing behind a transparent render leaves \BgIdentRate\% of top-down scenes byte-identical.
Lighting, focal length, and pitch can also change what a human can see, so their entire effect cannot be labeled evaluator noise.

\subsection{Prompt-side sensitivity: the template is the largest knob}
\label{sec:results-text}

For 17 of \NAlign{} alignment evaluators, the choice among 337 caption-template grammar paths moves the score more than any render factor in the default-conditioned sweep (Figure~\ref{fig:heatmap}).
The exceptions are BLIP-VQA, with pitch dispersion of $40\%$ against template dispersion of $32\%$, and the hosted judge, whose template $\rho$ stays below one at $0.87$ while pitch dominates.
For OpenAI CLIP ViT-L/14, mean per-scene normalized range is \ClipLTemplateDisp\% across the grammar and \ClipLTemplateReal\% even across common caption forms such as the bare prompt and \emph{a top-down view of}.
The scene and task prompt remain fixed, although wrappers can assert a viewpoint or image type that may be inaccurate.
The hosted GPT-4o judge has template ranges of \GptTemplateDisp\% and \GptTemplateReal\%, and Appendix~\ref{app:hosted-provenance} documents its zero-score handling and sensitivity bounds.
Paraphrase results are secondary because the rewrite audit found meaning drift in about half of the rewrites (Appendix~\ref{app:paraphrase-audit}).
Template choice is also directional: the best grammar path beats the default \emph{a top-down view of} by a median \BestTmplInfl\% (Appendix~\ref{app:template-effect}).
The pattern spans multiple CLIP variants: peak $\rho$ exceeds one for every alignment evaluator, with a minimum of $3.5$, and template dispersion and weak scramble responses persist across variants (Tables~\ref{tab:dispersion} and~\ref{tab:content}).
Table~\ref{tab:slotgrammar-effect} decomposes the wording effect by matched grammar slots.
The advantage of \emph{rendering} over \emph{photo} may partly reflect a truthful image-type description, so these shifts are protocol effects rather than uniformly evaluator defects.

\begin{table}[t]
    \caption{Per-slot caption-template effects over the \NounSweepN{} text-swept evaluators.
        Dispersion is the range of choice means as a percentage of the slot mean, and shifts are relative to that mean.
        Each slot is measured on a matched subset of captions to limit confounding by co-occurring slots.
        \emph{(none)} omits an optional slot.}
    \label{tab:slotgrammar-effect}
    \centering
    \small
    \begin{tabular}{@{}lll@{}}
        \toprule
        \textbf{Slot} & \textbf{Disp.} & \textbf{Choices} \\
        \midrule
        Viewpoint & $18\%$ & \emph{top-down}~$+6\%$, \emph{bird-eye}~$+5\%$, (none)~$-11\%$ \\
Head noun & $10\%$ & \emph{rendering}~$+6\%$, \emph{view}~$+1\%$, \emph{image}~$-1\%$, \emph{picture}~$-2\%$, \emph{photo}~$-4\%$ \\
Scene frame & $13\%$ & \emph{a virtual scene:}~$+8\%$, \emph{a scene:}~$-2\%$, \emph{scene:}~$-2\%$, (none)~$-4\%$ \\
Connective & $3\%$ & \emph{depicting}~$+1\%$, \emph{about}~$0\%$, \emph{of}~$0\%$ \\
Article & $4\%$ & \emph{a}~$+1\%$, (none)~$0\%$, \emph{an}~$-1\%$ \\
                 \\[-1.2em]
        \bottomrule
    \end{tabular}
\end{table}

\subsection{Score movement does not by itself establish a winner reversal}
\label{sec:results-rank}

Peak $\rho$ exceeds one for \NRhoAlignPeakGtOne/\NAlign{} alignment evaluators as a point estimate and retains a prompt-bootstrap lower bound above one for \NRhoAlignCiGtOne/\NAlign{} (Table~\ref{tab:rho-ci}).
Leaving out each generator in turn preserves the threshold for \NRhoAlignLogoGtOne{} alignment evaluators, but does not cover a wholly different generator field.
A common score shift cancels in pairwise margins, so $\rho>1$ alone cannot show a comparison change (Equation~\ref{eq:margin-shift}).

Aggregate rankings remain fairly concordant, with median minimum Kendall's $W=0.58$ across all 20 evaluators, but \AuditPointEstimateReversals{} of \NAlign{} alignment evaluators have more than one top-1 generator across factor levels.
At baseline, a median of \MedSigPairs{} of \NPairsTotal{} generator pairs are separable by paired prompt bootstrap, yet this confirms no reversal for any pair.
Under the same 50 prompt blocks, \AuditPointwiseReversals{} evaluators have opposite pointwise $95\%$ intervals at two levels of one factor, and none does so under the approximate simultaneous band over \AuditNContrasts{} searched contrasts (Appendix~\ref{app:selection-inference}).
The pointwise count changes from nine to eight under a second bootstrap seed, while the simultaneous count stays zero.
We therefore report changes in the point-estimate winner, not statistically confirmed reversals after full-search correction.
The absence of confirmation is not evidence that configurations have no effect.
A selected example shows pointwise intervals excluding zero in opposite directions while both simultaneous intervals include zero (Table~\ref{tab:reversal-examples}).

\begin{table}[t]
    \caption{Selected same-pair point-estimate reversals at two configurations.
        Gaps keep a fixed first-minus-second orientation within each row, and scales differ across evaluators.
        Pointwise $95\%$ intervals exclude zero in opposite directions.
        In every row at least one of the two full-search simultaneous intervals includes zero.
        The BLIP-VQA pointwise result changes under a second bootstrap seed.}
    \label{tab:reversal-examples}
    \centering
    \scriptsize
    \setlength{\tabcolsep}{2.5pt}
    \begin{tabular}{@{}lllll@{}}
        \toprule
        \textbf{Evaluator / factor} & \textbf{Method pair} & \textbf{Levels} & \textbf{Gap [pointwise 95\%]} & \textbf{Simultaneous 95\%} \\
        \midrule
\shortstack[l]{CLIP ViT-B/32\\Lighting} & Scenethesis $-$ I-Design & \shortstack[l]{city\\night} & \shortstack[l]{$+0.91\ [+0.12, +1.72]$\\$-0.91\ [-1.62, -0.20]$} & \shortstack[l]{$[-0.87, +2.68]$\\$[-2.54, +0.72]$} \\
\shortstack[l]{CLIP ViT-L/14\\Pitch} & HSM $-$ Reason-3D & \shortstack[l]{15$^{\circ}$\\90$^{\circ}$} & \shortstack[l]{$+1.49\ [+0.26, +2.74]$\\$-3.96\ [-5.08, -2.87]$} & \shortstack[l]{$[-1.27, +4.26]$\\$[-6.49, -1.44]$} \\
\shortstack[l]{CLIP ViT-L/14@336\\Pitch} & Scenethesis $-$ Reason-3D & \shortstack[l]{15$^{\circ}$\\90$^{\circ}$} & \shortstack[l]{$+1.53\ [+0.52, +2.62]$\\$-1.39\ [-2.57, -0.23]$} & \shortstack[l]{$[-0.78, +3.83]$\\$[-4.05, +1.27]$} \\
\shortstack[l]{CLIP ViT-L/14@336\\Pitch} & HSM $-$ Reason-3D & \shortstack[l]{30$^{\circ}$\\90$^{\circ}$} & \shortstack[l]{$+1.61\ [+0.62, +2.68]$\\$-5.11\ [-6.14, -4.06]$} & \shortstack[l]{$[-0.77, +3.99]$\\$[-7.49, -2.72]$} \\
\shortstack[l]{OpenCLIP ViT-bigG/14\\Pitch} & HSM $-$ LayoutVLM & \shortstack[l]{45$^{\circ}$\\90$^{\circ}$} & \shortstack[l]{$+2.26\ [+0.76, +3.66]$\\$-11.29\ [-13.47, -8.99]$} & \shortstack[l]{$[-0.97, +5.50]$\\$[-16.44, -6.14]$} \\
\shortstack[l]{MetaCLIP ViT-L/14\\Pitch} & Scenethesis $-$ LayoutVLM & \shortstack[l]{0$^{\circ}$\\75$^{\circ}$} & \shortstack[l]{$+2.31\ [+1.03, +3.67]$\\$-1.57\ [-2.88, -0.22]$} & \shortstack[l]{$[-0.61, +5.24]$\\$[-4.52, +1.39]$} \\
\shortstack[l]{BLIP-2 ITM\\Template} & Scenethesis $-$ HSM & \shortstack[l]{\texttt{T1}\\\texttt{T2}} & \shortstack[l]{$+7.37\ [+2.01, +13.75]$\\$-12.08\ [-23.57, -0.64]$} & \shortstack[l]{$[-6.07, +20.81]$\\$[-37.07, +12.90]$} \\
\shortstack[l]{BLIP-2 ITC\\Template} & Scenethesis $-$ HSM & \shortstack[l]{\texttt{T3}\\\texttt{T4}} & \shortstack[l]{$+2.45\ [+0.72, +4.23]$\\$-1.61\ [-2.92, -0.29]$} & \shortstack[l]{$[-1.51, +6.40]$\\$[-4.54, +1.31]$} \\
\shortstack[l]{BLIP-VQA\\Template} & Scenethesis $-$ Holodeck & \shortstack[l]{\texttt{T5}\\\texttt{T6}} & \shortstack[l]{$+4.33\ [+0.13, +8.93]$\\$-7.52\ [-11.81, -3.67]$} & \shortstack[l]{$[-5.56, +14.22]$\\$[-16.94, +1.90]$} \\
\shortstack[l]{VQAScore\\Pitch} & Scenethesis $-$ HSM & \shortstack[l]{75$^{\circ}$\\0$^{\circ}$} & \shortstack[l]{$+30.82\ [+24.28, +37.09]$\\$-5.83\ [-11.32, -0.69]$} & \shortstack[l]{$[+16.51, +45.14]$\\$[-17.74, +6.09]$} \\
                                                                                                  \\[-1.2em]
        \bottomrule
    \end{tabular}
    \par\smallskip
    \parbox{\linewidth}{\scriptsize \texttt{T1} \emph{Picture about a scene:}, \texttt{T2} \emph{Rendering about scene:}, \texttt{T3} \emph{A picture depicting scene:}, \texttt{T4} \emph{Bird-eye rendering about}, \texttt{T5} \emph{A top-down rendering about a virtual scene:}, and \texttt{T6} the bare prompt, each template ending with the prompt.}
    \vspace{-1.5em}
\end{table}

The margin envelopes provide a complementary descriptive result: 0--7 of 15 generator pairs per alignment evaluator retain a strict ordering across the baseline and 386 shared factor levels (Equation~\ref{eq:margin-envelope}, Table~\ref{tab:protocol-margins}).
The rest cross zero somewhere in this finite protocol set.
Only the hosted judge retains one winner throughout.
Re-executing the audited Holodeck protocol on the frozen scenes flips six of the 10 anchor pair margins that avoid Reason-3D's inverted pitch convention under at least one neighboring plausible configuration (Appendix~\ref{app:case-study}).
Stable orderings are neither simultaneous confidence statements nor evidence of human-judged correctness.

A preregistered $2\times2$ factorial crossing caption form with semantic pitch estimates a caption-by-pitch interaction of $-0.523$ $[-0.670,-0.383]$ on the pinned evaluator, so the top-down caption advantage shrinks at semantic pitch 45, while a supplementary hosted-judge run shows no interaction and a descriptive OpenAI ViT-L/14 run shows a much smaller one (Appendix~\ref{app:followup-factorial}).
In a preregistered comparison of aggregation rules at a fixed 15-pose render budget, multi-view means flip five of 15 generator pairs for one pinned evaluator and two of 15 for the other, and switching between the mean and the maximum flips one to three further pairs, so averaging is another declared protocol choice rather than a remedy (Appendix~\ref{app:followup-viewagg}).

\subsection{Content probes: weak targeted arrangement sensitivity}
\label{sec:results-content}

\begin{figure}[t]
    \centering
    \includegraphics[width=0.82\linewidth]{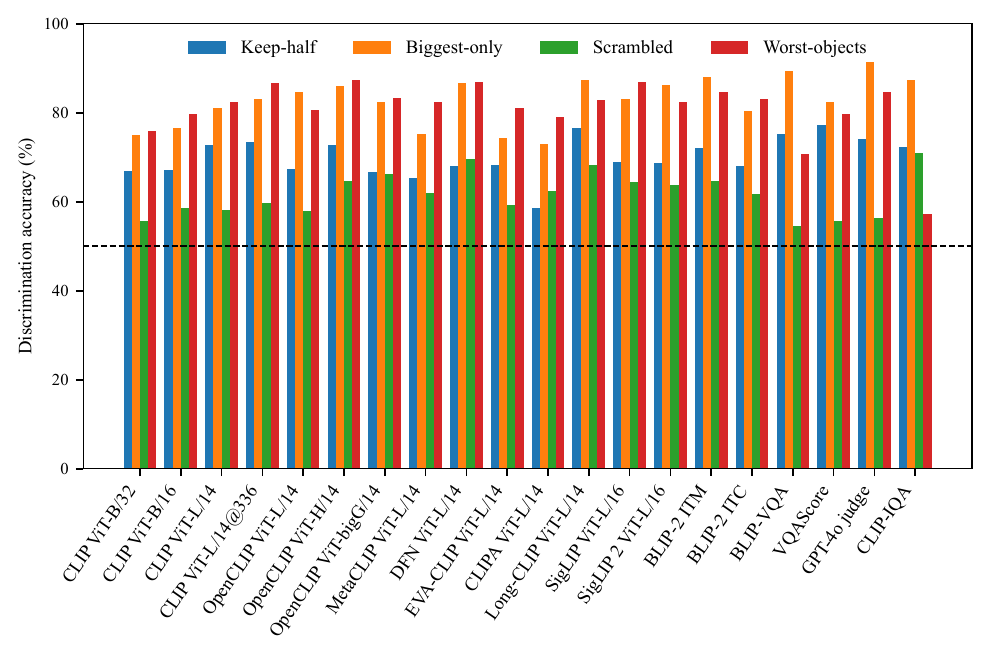}
    \vspace{-1.5em}
    \caption{Tie-adjusted directional discrimination against each intact scene, with the $50\%$ chance line dashed.
        Removing objects is easier to detect than scrambling their layout across this fixed dataset.
        The scramble ceiling belongs to the prompt-free CLIP-IQA control.}
    \label{fig:content}
    \vspace{-1.5em}
\end{figure}

Figure~\ref{fig:content} compares the four targeted content probes, with per-evaluator values in Table~\ref{tab:content}.
Removal probes are easier than arrangement: keeping only the largest object is detected at $73$--$92\%$ tie-adjusted directional discrimination.
For layout scrambling, no evaluator exceeds $71\%$ and most score $55$--$66\%$ against the intact scene.
The ceiling belongs to CLIP-IQA, a control that never reads the prompt and can react to visual plausibility rather than text-layout agreement.
For the hosted GPT-4o judge, \AuditGptZeroPairs{} of 300 intact--scrambled pairs contain a zero score.
Its tie-adjusted discrimination is \AuditGptCreditPct\%, and assigning arbitrary credit to those zero-score pairs yields sensitivity bounds of $[\AuditGptSensitivityLo,\AuditGptSensitivityHi]\%$ (Appendix~\ref{app:hosted-provenance}).
Scrambling is a necessary-condition probe, and the preregistered blinded study of Appendix~\ref{app:human-study} anchors it: raters detect the degradation at $82\%$ yet tie on 14 of 25 camera pairs whose metric margins are just as large.

\subsection{Best-view selection can inflate a reported score}
\label{sec:results-bestview}

Selecting the highest-scoring tested pitch/yaw leaves the scene fixed but raises median scores by $1.03$--$1.18\times$ for CLIP encoders, $1.29\times$ for VQAScore, and \GptBestView$\times$ for the GPT-4o judge (Table~\ref{tab:headline}).
Those figures measure selection over saved poses and are lower bounds on the maximum score available from a larger pose search.
The preferred poses are systematic: telephoto, bright, and shallow-pitch levels score highest for nearly every evaluator (Table~\ref{tab:bestlevel}).

\section{Discussion}
\label{sec:discussion}

\paragraph{Reliability is not validity.}
Score stability, ranking agreement, and targeted sensitivity answer different questions, and a high concordance over six generators can coexist with an unstable point-estimate winner when the leading methods are close.
Our simultaneous analysis does not establish reversed generator superiority, and the scrambling probe does not replace human judgments of scene quality.
Likewise, a one-signed protocol margin envelope establishes consistency of the sampled ordering across declared configurations, not correctness of that ordering.
The useful outcome is scoped evidence: which direction is stable, which depends on the protocol, and which remains statistically unresolved.

\paragraph{Fixed artifacts still yield different evidence.}
Changing the camera can reveal a relation, hide an object, or make a scene easier to inspect, and an alignment evaluator should respond when visible evidence changes, so identical scores across views are not the target.
The risk is drawing a scene-level conclusion from an unspecified or selectively chosen view.
Human judgments of camera-pair legibility and intact--scrambled alignment are therefore separate validation targets.
Caption wrappers also need semantic audit because a correct image-type assertion and an incorrect viewpoint assertion are different controls.

\paragraph{Report a contract before claiming superiority.}
Reporting \emph{(generator, score, card ID)} makes camera, caption, uncertainty, companion checks, and evidence status auditable.
With six generators and a one-factor-at-a-time core, the audit is a diagnostic and reporting procedure.
The blinded human comparison of Appendix~\ref{app:human-study} supplies one validation axis and classifies part of the configuration variance as metric artifact.

\paragraph{Preregistered follow-ups estimate what the audit could not.}
A $2\times2$ caption-by-pitch factorial over the \NumScenes{} frozen scenes gives an interaction of $-0.523$ with a $95\%$ prompt-bootstrap interval of $[-0.670,-0.383]$ on the evaluator pinned by a pre-declared rule, so the top-down caption advantage shrinks at the higher pitch level.
The hosted judge shows no interaction and a descriptive OpenAI ViT-L/14 run shows a much smaller one, so the interaction estimate is itself evaluator-dependent.
A preregistered comparison of view-aggregation rules at one fixed 15-pose budget finds the multi-view mean still flips five of 15 pair signs for one evaluator and two of 15 for the other, so averaging does not remove configuration dependence.

\paragraph{Which comparisons to trust.}
Under the same approximate simultaneous band, no claimed reversal survives, and exactly one strict ordering does, a loss for Holodeck.
Scenethesis over Holodeck holds for one evaluator and no other ordering survives, so no generator comparison is simultaneously confirmed across this protocol family.
Equivalence testing with a per-evaluator bound classifies $90.5\%$ of resolution shifts but only $37.9\%$ of pitch shifts as practically negligible, so resolution and background shifts are usually ignorable and pitch and yaw shifts are not.
We therefore recommend reporting protocol margins and the configuration card rather than single-view point winners.

\paragraph{How to use the audit.}
Fix the scene artifacts, identify the full evaluator protocol, and declare the plausible configuration set before inspecting outcomes.
Report pairwise margins over that set alongside the default score, since a margin envelope that changes sign calls for a configuration-dependent comparison.
Preserve prompt pairing, account for the searched comparison family, distinguish request failures from observed low scores, and add a property-matched or human-grounded sensitivity check.

\section{Conclusion}
\label{sec:conclusion}

Text-to-3D evaluation is not specified by naming a metric.
Across \NumScenes{} frozen scenes, configuration variance often exceeds between-generator variance, and descriptive winners can change even when aggregate rankings remain relatively stable, but those changes do not survive the full-search simultaneous analysis.
Pairwise protocol margin envelopes add the constructive result: some sampled comparisons keep their direction across the declared audit and others need configuration-dependent claims.
A credible comparison should report the evaluator configuration, uncertainty, stable comparison scope, and independent sensitivity evidence, tied together by \emph{(generator, score, card ID)}.

\subsection*{AI use statement}
Generative AI tools are part of the experimental pipeline: they produce the intended semantics-preserving paraphrase probes, serve as optional language backbones for applicable scene generators, and provide the hosted MLLM evaluator described in Section~\ref{sec:method}.
We also used generative AI tools for code and manuscript editing.
The rebuttal-phase follow-up experiments and analyses, including the factorial scoring, the aggregation comparison, the zero re-scoring, and the statistical reanalyses, were executed with AI assistance under author direction, and the authors verified the resulting artifacts and take responsibility for them.
The authors take responsibility for the final content of this work, including text, claims, code, and artifacts produced with the aid of generative AI.

\subsection*{Ethics statement}
This work audits automatic evaluators and the reported audit introduces no human participants or personally identifiable data.
The added blinded human comparison is a minimal-risk study with adult volunteer raters recruited from the authors' institution.
Participation is voluntary, with consent recorded in the survey before the first trial, and the median session lasts about half an hour.
The analyzed judgments are stored as anonymous pairwise records under a pseudonymous rater identifier.
The survey additionally offered an optional comment box and an optional contact field for the follow-up discussion, and both fields are excluded from the analysis and from every released record.
All prompts are drawn from publicly accessible sources.
The generated scenes are produced by existing systems and are used only as fixed inputs to the evaluators under study.
Our findings concern the reliability of evaluation metrics and are intended to improve measurement practice.
We see no direct pathway to harmful use.
The paper's claims concern evaluation protocols, not generator superiority: method-pair differences illustrate measurement sensitivity rather than establish which generator is best.

\subsection*{Reproducibility statement}
Regenerating the reported tables requires 20 raw evaluator files with 6,000 evaluator-scene records covering the same 300 scenes, plus structured descriptions of each intact scene and its four variants.
The repository provides validators for evaluator identity, scene coverage, prompt correspondence, factor levels, and deduplicated caption sets.
Its hash-bound intake-manifest contract is designed to bind the source revision and raw inputs to regenerated numbers.
Table~\ref{tab:evaluators} records evaluator identifiers, Table~\ref{tab:factors} records factor levels, and Appendix~\ref{app:templates} specifies the caption grammar.
For GPT scrambling, the same 300 score pairs yield 127 strict drops, 84 exact ties, and 89 rises, explaining the difference between a $42.3\%$ strict-drop rate and $56.3\%$ tie-adjusted discrimination.
The hosted records identify the model behind each hosted-judge score.
Freezing the saved scores fixes the analyzed evidence behind every reported number.
The revised supplementary archive contains the evaluation-harness source with its vendored LAVIS dependency, the complete frozen raw records for all twenty evaluators at both prompt sets, the verified intake manifest, and the follow-up evidence records.
All twenty SHA-256 hashes in the released manifest match the manifest committed to the repository, so every reported number can be recomputed from the frozen records.
The original renders and the full hosted request logs stay outside the archive because of their size and are committed as a hosted release at camera-ready.
Zero-score cells are handled by the deterministic sensitivity range of Appendix~\ref{app:hosted-provenance} and by the re-scoring analysis reported there.

\bibliographystyle{iclr2027_conference}
\bibliography{iclr2027_conference}

\begin{thebibliography}{64}
\providecommand{\natexlab}[1]{#1}
\providecommand{\url}[1]{\texttt{#1}}
\expandafter\ifx\csname urlstyle\endcsname\relax
  \providecommand{\doi}[1]{doi: #1}\else
  \providecommand{\doi}{doi: \begingroup \urlstyle{rm}\Url}\fi

\bibitem[Agarwal et~al.(2021)Agarwal, Schwarzer, Castro, Courville, and
  Bellemare]{rliable}
Rishabh Agarwal, Max Schwarzer, Pablo~Samuel Castro, Aaron Courville, and Marc
  Bellemare.
\newblock Deep reinforcement learning at the edge of the statistical precipice.
\newblock In Marc'Aurelio Ranzato, Alina Beygelzimer, Yann Dauphin, Percy
  Liang, and Jennifer Wortman~Vaughan (eds.), \emph{Advances in Neural
  Information Processing Systems (NeurIPS)}, volume~34, pp.\  29304--29320,
  2021.

\bibitem[Asaria et~al.(2026)Asaria, Salomone, and Gandhi]{asaria2026crossmodel}
Ali Asaria, Tony Salomone, and Deep Gandhi.
\newblock A cross-model {VLM}-judge protocol for single-image {3D} mesh quality
  (and why cheap proxies fall short).
\newblock \emph{arXiv preprint arXiv:2606.18451}, 2026.

\bibitem[Bai et~al.(2025)Bai, Bai, Chen, Wu, Li, and Ma]{freescene}
Tongyuan Bai, Wangyuanfan Bai, Dong Chen, Tieru Wu, Manyi Li, and Rui Ma.
\newblock { FreeScene: Mixed Graph Diffusion for 3D Scene Synthesis from Free
  Prompts }.
\newblock In \emph{IEEE/CVF Conference on Computer Vision and Pattern
  Recognition (CVPR)}, pp.\  5893--5903, Los Alamitos, CA, USA, June 2025.
  IEEE.
\newblock \doi{10.1109/CVPR52734.2025.00553}.

\bibitem[Berdoz et~al.(2026)Berdoz, Lanzend\"{o}rfer, Tuninga, and
  Wattenhofer]{reason3d}
Fr\'{e}d\'{e}ric Berdoz, Luca~A. Lanzend\"{o}rfer, Nick Tuninga, and Roger
  Wattenhofer.
\newblock Text-to-scene with large reasoning models.
\newblock In \emph{AAAI Conference on Artificial Intelligence}, volume~40, pp.\
   2435--2443. AAAI Press, 2026.
\newblock ISBN 978-1-57735-906-7.
\newblock \doi{10.1609/aaai.v40i4.37229}.

\bibitem[Bian et~al.(2025)Bian, Ren, Yang, and Callison-Burch]{holodeck2}
Zixuan Bian, Ruohan Ren, Yue Yang, and Chris Callison-Burch.
\newblock {HOLODECK} 2.0: Vision-language-guided {3D} world generation with
  editing.
\newblock \emph{arXiv preprint arXiv:2508.05899}, 2025.

\bibitem[Bordes et~al.(2025)Bordes, Ross, Kao, Spiliopoulou, and
  Williams]{bordes2025evalfactsheets}
Florian Bordes, Candace Ross, Justine~T. Kao, Evangelia Spiliopoulou, and Adina
  Williams.
\newblock Eval factsheets: A structured framework for documenting {AI}
  evaluations.
\newblock \emph{arXiv preprint arXiv:2512.04062}, 2025.

\bibitem[Burapacheep et~al.(2024)Burapacheep, Gaur, Bhatia, and
  Thrush]{colorswap}
Jirayu Burapacheep, Ishan Gaur, Agam Bhatia, and Tristan Thrush.
\newblock {C}olor{S}wap: A color and word order dataset for multimodal
  evaluation.
\newblock In Lun-Wei Ku, Andre Martins, and Vivek Srikumar (eds.),
  \emph{Findings of the Association for Computational Linguistics: ACL 2024},
  pp.\  1716--1726, Bangkok, Thailand, August 2024. Association for
  Computational Linguistics.
\newblock \doi{10.18653/v1/2024.findings-acl.99}.

\bibitem[\c{C}elen et~al.(2024)\c{C}elen, Han, Schindler, Van~Gool, Armeni,
  Obukhov, and Wang]{idesign}
Ata \c{C}elen, Guo Han, Konrad Schindler, Luc Van~Gool, Iro Armeni, Anton
  Obukhov, and Xi~Wang.
\newblock {I-Design}: Personalized {LLM} interior designer.
\newblock In \emph{European Conference on Computer Vision (ECCV) Workshops},
  volume 15624 of \emph{Lecture Notes in Computer Science}, pp.\  217--234,
  Cham, 2024. Springer Nature Switzerland.
\newblock ISBN 978-3-031-92386-9.
\newblock \doi{10.1007/978-3-031-92387-6_17}.

\bibitem[Charles et~al.(2025)Charles, Panda, Agarwal, Patel, Pattnayak, Kumar,
  and Kumar]{diagnosticaudit}
Angeline Charles, Srikant Panda, Amit Agarwal, Hitesh~Laxmichand Patel,
  Priyaranjan Pattnayak, Bhargava Kumar, and Tejaswini Kumar.
\newblock A diagnostic framework for auditing reference-free vision-language
  metrics.
\newblock In Kentaro Inui, Sakriani Sakti, Haofen Wang, Derek~F. Wong, Pushpak
  Bhattacharyya, Biplab Banerjee, Asif Ekbal, Tanmoy Chakraborty, and
  Dhirendra~Pratap Singh (eds.), \emph{International Joint Conference on
  Natural Language Processing and the 4th Conference of the Asia-Pacific
  Chapter of the Association for Computational Linguistics (IJCNLP-AACL)}, pp.\
   2633--2644, Mumbai, India, December 2025. The Asian Federation of Natural
  Language Processing and The Association for Computational Linguistics.
\newblock ISBN 979-8-89176-298-5.
\newblock \doi{10.18653/v1/2025.ijcnlp-long.142}.

\bibitem[Cherti et~al.(2023)Cherti, Beaumont, Wightman, Wortsman, Ilharco,
  Gordon, Schuhmann, Schmidt, and Jitsev]{cherti2023reproducible}
Mehdi Cherti, Romain Beaumont, Ross Wightman, Mitchell Wortsman, Gabriel
  Ilharco, Cade Gordon, Christoph Schuhmann, Ludwig Schmidt, and Jenia Jitsev.
\newblock Reproducible scaling laws for contrastive language-image learning.
\newblock In \emph{IEEE/CVF Conference on Computer Vision and Pattern
  Recognition (CVPR)}, pp.\  2818--2829, 2023.
\newblock \doi{10.1109/CVPR52729.2023.00276}.

\bibitem[Cho et~al.(2024)Cho, Hu, Garg, Anderson, Krishna, Baldridge, Bansal,
  Pont-Tuset, and Wang]{dsg}
Jaemin Cho, Yushi Hu, Roopal Garg, Peter Anderson, Ranjay Krishna, Jason
  Baldridge, Mohit Bansal, Jordi Pont-Tuset, and Su~Wang.
\newblock Davidsonian scene graph: Improving reliability in fine-grained
  evaluation for text-to-image generation.
\newblock In \emph{International Conference on Learning Representations
  (ICLR)}, volume 2024, pp.\  15625--15645, 2024.

\bibitem[Deng et~al.(2025)Deng, Qi, and Ma]{gltreesearch}
Wei Deng, Mengshi Qi, and Huadong Ma.
\newblock { Global-Local Tree Search in VLMs for 3D Indoor Scene Generation }.
\newblock In \emph{IEEE/CVF Conference on Computer Vision and Pattern
  Recognition (CVPR)}, pp.\  8975--8984, June 2025.
\newblock \doi{10.1109/CVPR52734.2025.00839}.

\bibitem[Duan et~al.(2025)Duan, Yu, Chen, Fei-Fei, and Wu]{worldscore}
Haoyi Duan, Hong-Xing Yu, Sirui Chen, Li~Fei-Fei, and Jiajun Wu.
\newblock {WorldScore}: A unified evaluation benchmark for world generation.
\newblock In \emph{IEEE/CVF International Conference on Computer Vision
  (ICCV)}, pp.\  27713--27724, October 2025.
\newblock \doi{10.1109/ICCV51701.2025.02573}.

\bibitem[Duggal et~al.(2025)Duggal, Hu, Michel, Kembhavi, Freeman, Smith,
  Krishna, Torralba, Farhadi, and Ma]{eval3d}
Shivam Duggal, Yushi Hu, Oscar Michel, Aniruddha Kembhavi, William~T. Freeman,
  Noah~A. Smith, Ranjay Krishna, Antonio Torralba, Ali Farhadi, and Wei-Chiu
  Ma.
\newblock {Eval3D}: Interpretable and fine-grained evaluation for {3D}
  generation.
\newblock In \emph{IEEE/CVF Conference on Computer Vision and Pattern
  Recognition (CVPR)}, pp.\  13326--13336, 2025.
\newblock \doi{10.1109/CVPR52734.2025.01244}.

\bibitem[Fang et~al.(2024)Fang, Madappally~Jose, Jain, Schmidt, Toshev, and
  Shankar]{dfn}
Alex Fang, Albin Madappally~Jose, Amit Jain, Ludwig Schmidt, Alexander Toshev,
  and Vaishaal Shankar.
\newblock Data filtering networks.
\newblock In \emph{International Conference on Learning Representations
  (ICLR)}, volume 2024, pp.\  36221--36237, 2024.

\bibitem[Ghosh et~al.(2026)Ghosh, Reuel, Chim, Kennedy, Yadav, Mickel, Long,
  Tran, et~al.]{evalcards}
Avijit Ghosh, Anka Reuel, Jenny Chim, Wm.~Matthew Kennedy, Srishti Yadav,
  Jennifer Mickel, Yanan Long, Andrew Tran, et~al.
\newblock Evaluation cards: An interpretive layer for {AI} evaluation
  reporting.
\newblock \emph{arXiv preprint arXiv:2606.09809}, 2026.

\bibitem[Gu et~al.(2025)Gu, Cui, Li, Wei, Ge, Gu, Liu, Davis, and
  Ding]{artiscene}
Zeqi Gu, Yin Cui, Zhaoshuo Li, Fangyin Wei, Yunhao Ge, Jinwei Gu, Ming-Yu Liu,
  Abe Davis, and Yifan Ding.
\newblock {ArtiScene}: Language-driven artistic {3D} scene generation through
  image intermediary.
\newblock In \emph{IEEE/CVF Conference on Computer Vision and Pattern
  Recognition (CVPR)}, pp.\  2891--2901, 2025.
\newblock \doi{10.1109/CVPR52734.2025.00275}.

\bibitem[He et~al.(2023)He, Bai, Lin, Zhao, Hu, Sheng, Yi, Li, and
  Liu]{t3bench}
Yuze He, Yushi Bai, Matthieu Lin, Wang Zhao, Yubin Hu, Jenny Sheng, Ran Yi,
  Juanzi Li, and Yong-Jin Liu.
\newblock {T$^3$Bench}: Benchmarking current progress in text-to-3d generation.
\newblock \emph{arXiv preprint arXiv:2310.02977}, 2023.

\bibitem[Hessel et~al.(2021)Hessel, Holtzman, Forbes, Le~Bras, and
  Choi]{clipscore}
Jack Hessel, Ari Holtzman, Maxwell Forbes, Ronan Le~Bras, and Yejin Choi.
\newblock {CLIPS}core: A reference-free evaluation metric for image captioning.
\newblock In Marie-Francine Moens, Xuanjing Huang, Lucia Specia, and Scott
  Wen-tau Yih (eds.), \emph{Conference on Empirical Methods in Natural Language
  Processing (EMNLP)}, pp.\  7514--7528, Online and Punta Cana, Dominican
  Republic, November 2021. Association for Computational Linguistics.
\newblock \doi{10.18653/v1/2021.emnlp-main.595}.

\bibitem[Hou et~al.(2025)Hou, Zhao, Xu, Fan, and Jiang]{tdbench}
Kaiyuan Hou, Minghui Zhao, Lilin Xu, Yuang Fan, and Xiaofan Jiang.
\newblock {TDBench}: A benchmark for top-down image understanding with
  reliability analysis of vision-language models.
\newblock \emph{arXiv preprint arXiv:2504.03748}, 2025.

\bibitem[Hsieh et~al.(2023)Hsieh, Zhang, Ma, Kembhavi, and Krishna]{sugarcrepe}
Cheng-Yu Hsieh, Jieyu Zhang, Zixian Ma, Aniruddha Kembhavi, and Ranjay Krishna.
\newblock {SugarCrepe}: Fixing hackable benchmarks for vision-language
  compositionality.
\newblock In \emph{Advances in Neural Information Processing Systems (NeurIPS)
  Datasets and Benchmarks Track}, volume~36, pp.\  31096--31116, 2023.
\newblock \doi{10.52202/075280-1355}.

\bibitem[Hu et~al.(2023)Hu, Liu, Kasai, Wang, Ostendorf, Krishna, and
  Smith]{tifa}
Yushi Hu, Benlin Liu, Jungo Kasai, Yizhong Wang, Mari Ostendorf, Ranjay
  Krishna, and Noah~A. Smith.
\newblock {TIFA}: Accurate and interpretable text-to-image faithfulness
  evaluation with question answering.
\newblock In \emph{IEEE/CVF International Conference on Computer Vision
  (ICCV)}, pp.\  20349--20360, 2023.
\newblock \doi{10.1109/ICCV51070.2023.01866}.

\bibitem[Huang et~al.(2023)Huang, Sun, Xie, Li, and Liu]{t2icompbench}
Kaiyi Huang, Kaiyue Sun, Enze Xie, Zhenguo Li, and Xihui Liu.
\newblock {T2I-CompBench}: A comprehensive benchmark for open-world
  compositional text-to-image generation.
\newblock In \emph{Advances in Neural Information Processing Systems
  (NeurIPS)}, volume~36, pp.\  78723--78747, 2023.
\newblock \doi{10.52202/075280-3443}.

\bibitem[Huang et~al.(2025)Huang, Zhai, Bauer, Pollefeys, Tombari, Guibas,
  Huang, and Engelmann]{fpvscene}
Rui Huang, Guangyao Zhai, Zuria Bauer, Marc Pollefeys, Federico Tombari,
  Leonidas Guibas, Gao Huang, and Francis Engelmann.
\newblock Video perception models for 3d scene synthesis.
\newblock In \emph{Advances in Neural Information Processing Systems
  (NeurIPS)}, volume~38, pp.\  112687--112715, 2025.
\newblock \doi{10.52202/085713-3763}.

\bibitem[Hwang et~al.(2025)Hwang, Lee, Min, Kang, Kim, and Jung]{foolinglvlm}
Yerin Hwang, Dongryeol Lee, Kyungmin Min, Taegwan Kang, Yongil Kim, and Kyomin
  Jung.
\newblock Fooling the {LVLM} judges: Visual biases in {LVLM}-based evaluation.
\newblock In \emph{Conference on Empirical Methods in Natural Language
  Processing (EMNLP)}, pp.\  23186--23205. Association for Computational
  Linguistics, 2025.
\newblock \doi{10.18653/v1/2025.emnlp-main.1182}.

\bibitem[Jia et~al.(2026)Jia, Zhao, Huang, Wang, and Zheng]{db3dme}
Nanshan Jia, Zhenyu Zhao, Sui Huang, Jingshen Wang, and Zeyu Zheng.
\newblock {DB-3DME}: From dataset to benchmark for human-aligned automatic {3D}
  mesh evaluation.
\newblock In \emph{IEEE/CVF Conference on Computer Vision and Pattern
  Recognition (CVPR) Workshops}, pp.\  4631--4640, 2026.

\bibitem[Kang et~al.(2026)Kang, Choi, Hwangbo, Lim, Ong, and Yeo]{legoeval}
Minseok Kang, Dongwook Choi, Gyeom Hwangbo, Seungwon Lim, Kai Tzu-iunn Ong, and
  Jinyoung Yeo.
\newblock Apples on the table? evaluating text-guided {3D} scene synthesis via
  fine-grained constraint verification.
\newblock In \emph{Conference on Empirical Methods in Natural Language
  Processing (EMNLP)}, 2026.

\bibitem[Khan et~al.(2026)Khan, Suryanarayanan, Anand, and
  Khapra]{seeingbelieving}
Mohammed Safi Ur~Rahman Khan, Sanjay Suryanarayanan, Tushar Anand, and Mitesh~M
  Khapra.
\newblock Seeing isn't believing: Uncovering blind spots in evaluator
  vision-language models.
\newblock \emph{arXiv preprint arXiv:2604.21523}, 2026.

\bibitem[Kumar et~al.(2026)Kumar, Tayebati, Naik, Krishnan, and
  Trivedi]{vlmrankscore}
Divake Kumar, Sina Tayebati, Devashri Naik, Ranganath Krishnan, and Amit~Ranjan
  Trivedi.
\newblock {VLM} judges can rank but cannot score: Task-dependent uncertainty in
  multimodal evaluation.
\newblock \emph{arXiv preprint arXiv:2604.25235}, 2026.

\bibitem[Li et~al.(2023{\natexlab{a}})Li, Li, Savarese, and Hoi]{blip2}
Junnan Li, Dongxu Li, Silvio Savarese, and Steven Hoi.
\newblock {BLIP-2}: Bootstrapping language-image pre-training with frozen image
  encoders and large language models.
\newblock In \emph{International Conference on Machine Learning (ICML)}, volume
  202 of \emph{Proceedings of Machine Learning Research}, pp.\  19730--19742.
  PMLR, 2023{\natexlab{a}}.

\bibitem[Li et~al.(2026{\natexlab{a}})Li, Qu, Zhang, Gui, Xu, Zhang, Hu, Wang,
  and Wang]{genarena}
Ruihang Li, Leigang Qu, Jingxu Zhang, Dongnan Gui, Mengde Xu, Xiaosong Zhang,
  Han Hu, Wenjie Wang, and Jiaqi Wang.
\newblock {GenArena}: How can we achieve human-aligned evaluation for visual
  generation tasks?
\newblock \emph{arXiv preprint arXiv:2602.06013}, 2026{\natexlab{a}}.

\bibitem[Li et~al.(2026{\natexlab{b}})Li, Lam, Peng, Wang, and
  Lyu]{constraintir}
Shuqing Li, Anson~Y. Lam, Yun Peng, Wenxuan Wang, and Michael~R. Lyu.
\newblock {3D} software synthesis driven by constraint-expressive intermediate
  representation.
\newblock In \emph{IEEE/ACM International Conference on Software Engineering
  (ICSE)}, ICSE '26, pp.\  1354--1366, New York, NY, USA, 2026{\natexlab{b}}.
  Association for Computing Machinery.
\newblock ISBN 9798400720253.
\newblock \doi{10.1145/3744916.3787812}.

\bibitem[Li et~al.(2023{\natexlab{b}})Li, Wang, and Xie]{clipa}
Xianhang Li, Zeyu Wang, and Cihang Xie.
\newblock An inverse scaling law for {CLIP} training.
\newblock In \emph{Advances in Neural Information Processing Systems
  (NeurIPS)}, volume~36, pp.\  49068--49087, 2023{\natexlab{b}}.
\newblock \doi{10.52202/075280-2132}.

\bibitem[Lin \& Mu(2024)Lin and Mu]{instructscene}
Chenguo Lin and Yadong Mu.
\newblock {InstructScene}: Instruction-driven {3D} indoor scene synthesis with
  semantic graph prior.
\newblock In \emph{International Conference on Learning Representations
  (ICLR)}, 2024.

\bibitem[Lin et~al.(2024)Lin, Pathak, Li, Li, Xia, Neubig, Zhang, and
  Ramanan]{vqascore}
Zhiqiu Lin, Deepak Pathak, Baiqi Li, Jiayao Li, Xide Xia, Graham Neubig,
  Pengchuan Zhang, and Deva Ramanan.
\newblock Evaluating text-to-visual generation with image-to-text generation.
\newblock In \emph{European Conference on Computer Vision (ECCV)}, volume 15067
  of \emph{Lecture Notes in Computer Science}, pp.\  366--384, Cham, 2024.
  Springer Nature Switzerland.
\newblock ISBN 978-3-031-72672-9.
\newblock \doi{10.1007/978-3-031-72673-6_20}.

\bibitem[Michalkiewicz et~al.(2025)Michalkiewicz, Bai, Baktashmotlagh, Jampani,
  and Balakrishnan]{viewsnotequal}
Mateusz Michalkiewicz, Sheena Bai, Mahsa Baktashmotlagh, Varun Jampani, and
  Guha Balakrishnan.
\newblock Not all views are created equal: Analyzing viewpoint instabilities in
  vision foundation models.
\newblock In \emph{IEEE/CVF International Conference on Computer Vision
  (ICCV)}, pp.\  9113--9123, 2025.
\newblock \doi{10.1109/ICCV51701.2025.00852}.

\bibitem[Paiss et~al.(2023)Paiss, Ephrat, Tov, Zada, Mosseri, Irani, and
  Dekel]{clipcount10}
Roni Paiss, Ariel Ephrat, Omer Tov, Shiran Zada, Inbar Mosseri, Michal Irani,
  and Tali Dekel.
\newblock Teaching {CLIP} to count to ten.
\newblock In \emph{IEEE/CVF International Conference on Computer Vision
  (ICCV)}, pp.\  3147--3157, 2023.
\newblock \doi{10.1109/ICCV51070.2023.00294}.

\bibitem[Pun et~al.(2026)Pun, Tam, Wang, Huo, Chang, and Savva]{hsm}
Hou In~Derek Pun, Hou In~Ivan Tam, Austin~T Wang, Xiaoliang Huo, Angel~X Chang,
  and Manolis Savva.
\newblock {HSM}: Hierarchical scene motifs for multi-scale indoor scene
  generation.
\newblock In \emph{International Conference on 3D Vision (3DV)}, pp.\
  1356--1367. IEEE, 2026.

\bibitem[Radford et~al.(2021)Radford, Kim, Hallacy, Ramesh, Goh, Agarwal,
  Sastry, Askell, Mishkin, Clark, Krueger, and Sutskever]{clip}
Alec Radford, Jong~Wook Kim, Chris Hallacy, Aditya Ramesh, Gabriel Goh,
  Sandhini Agarwal, Girish Sastry, Amanda Askell, Pamela Mishkin, Jack Clark,
  Gretchen Krueger, and Ilya Sutskever.
\newblock Learning transferable visual models from natural language
  supervision.
\newblock In \emph{International Conference on Machine Learning (ICML)}, volume
  139, pp.\  8748--8763. PMLR, 2021.

\bibitem[Ross et~al.(2024)Ross, Hall, Romero~Soriano, and
  Williams]{whatmakesgoodmetric}
Candace Ross, Melissa Hall, Adriana Romero~Soriano, and Adina Williams.
\newblock What makes a good metric? evaluating automatic metrics for
  text-to-image consistency.
\newblock In \emph{Conference on Language Modeling (COLM)}, 2024.

\bibitem[Ryan et~al.(2026)Ryan, Zhang, Salunkhe, Chu, Xu, and
  Yang]{autometrics}
Michael~J. Ryan, Yanzhe Zhang, Amol Salunkhe, Yi~Chu, Di~Xu, and Diyi Yang.
\newblock {AutoMetrics}: Approximate human judgments with automatically
  generated evaluators.
\newblock In \emph{International Conference on Learning Representations
  (ICLR)}, 2026.

\bibitem[Saxon et~al.(2024)Saxon, Jahara, Khoshnoodi, Lu, Sharma, and
  Wang]{t2iscorescore}
Michael Saxon, Fatima Jahara, Mahsa Khoshnoodi, Yujie Lu, Aditya Sharma, and
  William~Yang Wang.
\newblock Who evaluates the evaluations? objectively scoring text-to-image
  prompt coherence metrics with {T2IScoreScore} ({TS2}).
\newblock In \emph{Advances in Neural Information Processing Systems
  (NeurIPS)}, volume~37, pp.\  85630--85657, 2024.
\newblock \doi{10.52202/079017-2719}.

\bibitem[Schaeffer et~al.(2023)Schaeffer, Miranda, and Koyejo]{metricmirage}
Rylan Schaeffer, Brando Miranda, and Sanmi Koyejo.
\newblock Are emergent abilities of large language models a mirage?
\newblock In \emph{Advances in Neural Information Processing Systems
  (NeurIPS)}, volume~36, pp.\  55565--55581, 2023.
\newblock \doi{10.52202/075280-2425}.

\bibitem[Sengupta et~al.(2026)Sengupta, Ao, and Cascante-Bonilla]{scenecritic}
Kathakoli Sengupta, Kai Ao, and Paola Cascante-Bonilla.
\newblock {SceneCritic}: A symbolic evaluator for {3D} indoor scene synthesis.
\newblock \emph{arXiv preprint arXiv:2604.13035}, 2026.
\newblock Poster at the 6th OpenSUN3D Workshop, CVPR 2026.

\bibitem[Sokol et~al.(2025)Sokol, Daly, Hind, Piorkowski, Zhang, Moniz, and
  Chawla]{sokol2025benchmarkcards}
Anna Sokol, Elizabeth Daly, Michael Hind, David Piorkowski, Xiangliang Zhang,
  Nuno Moniz, and Nitesh~V. Chawla.
\newblock {BenchmarkCards}: Standardized documentation for large language model
  benchmarks.
\newblock In \emph{Advances in Neural Information Processing Systems
  (NeurIPS)}, volume~38, pp.\  91271--91289, 2025.
\newblock \doi{10.52202/085713-2748}.

\bibitem[Sun et~al.(2025)Sun, Liu, Gu, Lim, Bhat, Tombari, Li, Haber, and
  Wu]{layoutvlm}
Fan-Yun Sun, Weiyu Liu, Siyi Gu, Dylan Lim, Goutam Bhat, Federico Tombari,
  Manling Li, Nick Haber, and Jiajun Wu.
\newblock { LayoutVLM: Differentiable Optimization of 3D Layout via
  Vision-Language Models }.
\newblock In \emph{IEEE/CVF Conference on Computer Vision and Pattern
  Recognition (CVPR)}, pp.\  29469--29478, Los Alamitos, CA, USA, June 2025.
  IEEE.
\newblock \doi{10.1109/CVPR52734.2025.02744}.

\bibitem[Sun et~al.(2023)Sun, Fang, Wu, Wang, and Cao]{evaclip}
Quan Sun, Yuxin Fang, Ledell Wu, Xinlong Wang, and Yue Cao.
\newblock {EVA-CLIP}: Improved training techniques for {CLIP} at scale.
\newblock \emph{arXiv preprint arXiv:2303.15389}, 2023.

\bibitem[Tam et~al.(2026)Tam, Pun, Wang, Chang, and Savva]{sceneeval}
Hou In~Ivan Tam, Hou In~Derek Pun, Austin~T. Wang, Angel~X. Chang, and Manolis
  Savva.
\newblock {SceneEval}: Evaluating semantic coherence in text-conditioned {3D}
  indoor scene synthesis.
\newblock In \emph{IEEE/CVF Winter Conference on Applications of Computer
  Vision (WACV)}, pp.\  7355--7365, 2026.
\newblock \doi{10.1109/WACV61042.2026.00710}.

\bibitem[Tang et~al.(2024)Tang, Nie, Markhasin, Dai, Thies, and
  Nie{\ss}ner]{diffuscene}
Jiapeng Tang, Yinyu Nie, Lev Markhasin, Angela Dai, Justus Thies, and Matthias
  Nie{\ss}ner.
\newblock {DiffuScene}: Denoising diffusion models for generative indoor scene
  synthesis.
\newblock In \emph{IEEE/CVF Conference on Computer Vision and Pattern
  Recognition (CVPR)}, pp.\  20507--20518, 2024.
\newblock \doi{10.1109/CVPR52733.2024.01938}.

\bibitem[Tschannen et~al.(2025)Tschannen, Gritsenko, Wang, Naeem,
  Alabdulmohsin, Parthasarathy, Evans, Beyer, Xia, Mustafa, et~al.]{siglip2}
Michael Tschannen, Alexey Gritsenko, Xiao Wang, Muhammad~Ferjad Naeem, Ibrahim
  Alabdulmohsin, Nikhil Parthasarathy, Talfan Evans, Lucas Beyer, Ye~Xia, Basil
  Mustafa, et~al.
\newblock {SigLIP} 2: Multilingual vision-language encoders with improved
  semantic understanding, localization, and dense features.
\newblock \emph{arXiv preprint arXiv:2502.14786}, 2025.

\bibitem[Wang et~al.(2023)Wang, Chan, and Loy]{clipiqa}
Jianyi Wang, Kelvin~C.K. Chan, and Chen~Change Loy.
\newblock Exploring {CLIP} for assessing the look and feel of images.
\newblock In \emph{AAAI Conference on Artificial Intelligence}, volume~37, pp.\
   2555--2563. AAAI Press, 2023.
\newblock ISBN 978-1-57735-880-0.
\newblock \doi{10.1609/aaai.v37i2.25353}.

\bibitem[Wiles et~al.(2025)Wiles, Zhang, Albuquerque, Kaji{\'c}, Wang,
  Bugliarello, Onoe, Papalampidi, Ktena, Knutsen, et~al.]{gecko}
Olivia Wiles, Chuhan Zhang, Isabela Albuquerque, Ivana Kaji{\'c}, Su~Wang,
  Emanuele Bugliarello, Yasumasa Onoe, Pinelopi Papalampidi, Ira Ktena,
  Christopher Knutsen, et~al.
\newblock Revisiting text-to-image evaluation with {Gecko}: On metrics,
  prompts, and human rating.
\newblock In \emph{International Conference on Learning Representations
  (ICLR)}, volume 2025, pp.\  272--287, 2025.

\bibitem[Wu et~al.(2024)Wu, Yang, Li, Zhang, Liu, Guibas, Lin, and
  Wetzstein]{gpteval3d}
Tong Wu, Guandao Yang, Zhibing Li, Kai Zhang, Ziwei Liu, Leonidas Guibas, Dahua
  Lin, and Gordon Wetzstein.
\newblock {GPT-4V(ision)} is a human-aligned evaluator for text-to-{3D}
  generation.
\newblock In \emph{IEEE/CVF Conference on Computer Vision and Pattern
  Recognition (CVPR)}, pp.\  22227--22238, 2024.
\newblock \doi{10.1109/CVPR52733.2024.02098}.

\bibitem[Xu et~al.(2024)Xu, Xie, Tan, Huang, Howes, Sharma, Li, Ghosh,
  Zettlemoyer, and Feichtenhofer]{metaclip}
Hu~Xu, Saining Xie, Xiaoqing Tan, Po-Yao Huang, Russell Howes, Vasu Sharma,
  Shang-Wen Li, Gargi Ghosh, Luke Zettlemoyer, and Christoph Feichtenhofer.
\newblock Demystifying {CLIP} data.
\newblock In \emph{International Conference on Learning Representations
  (ICLR)}, volume 2024, pp.\  47812--47831, 2024.

\bibitem[Yang et~al.(2025)Yang, Jia, Zhang, and Huang]{sceneweaver}
Yandan Yang, Baoxiong Jia, Shujie Zhang, and Siyuan Huang.
\newblock {SceneWeaver}: All-in-one {3D} scene synthesis with an extensible and
  self-reflective agent.
\newblock In \emph{Advances in Neural Information Processing Systems
  (NeurIPS)}, volume~38, pp.\  155685--155717, 2025.
\newblock \doi{10.52202/085713-4689}.

\bibitem[Yang et~al.(2024)Yang, Sun, Weihs, Vanderbilt, Herrasti, Han, Wu,
  Haber, Krishna, Liu, Callison-Burch, Yatskar, Kembhavi, and Clark]{holodeck}
Yue Yang, Fan-Yun Sun, Luca Weihs, Eli Vanderbilt, Alvaro Herrasti, Winson Han,
  Jiajun Wu, Nick Haber, Ranjay Krishna, Lingjie Liu, Chris Callison-Burch,
  Mark Yatskar, Aniruddha Kembhavi, and Christopher Clark.
\newblock Holodeck: Language guided generation of 3d embodied ai environments.
\newblock In \emph{IEEE/CVF Conference on Computer Vision and Pattern
  Recognition (CVPR)}, pp.\  16277--16287, 2024.
\newblock \doi{10.1109/CVPR52733.2024.01536}.

\bibitem[Ye et~al.(2024)Ye, Zheng, Liu, and Peng]{relscene}
Zhaoda Ye, Xinhan Zheng, Yang Liu, and Yuxin Peng.
\newblock {RelScene}: A benchmark and baseline for spatial relations in
  text-driven {3D} scene generation.
\newblock In \emph{ACM International Conference on Multimedia (ACM MM)}, pp.\
  10563--10571, 2024.
\newblock \doi{10.1145/3664647.3681653}.

\bibitem[Yuksekgonul et~al.(2023)Yuksekgonul, Bianchi, Kalluri, Jurafsky, and
  Zou]{aro}
Mert Yuksekgonul, Federico Bianchi, Pratyusha Kalluri, Dan Jurafsky, and James
  Zou.
\newblock When and why vision-language models behave like bags-of-words, and
  what to do about it?
\newblock In \emph{International Conference on Learning Representations
  (ICLR)}, 2023.

\bibitem[Zhai et~al.(2023)Zhai, Mustafa, Kolesnikov, and Beyer]{siglip}
Xiaohua Zhai, Basil Mustafa, Alexander Kolesnikov, and Lucas Beyer.
\newblock Sigmoid loss for language image pre-training.
\newblock In \emph{IEEE/CVF International Conference on Computer Vision
  (ICCV)}, pp.\  11941--11952, 2023.
\newblock \doi{10.1109/ICCV51070.2023.01100}.

\bibitem[Zhang et~al.(2024)Zhang, Zhang, Dong, Zang, and Wang]{longclip}
Beichen Zhang, Pan Zhang, Xiaoyi Dong, Yuhang Zang, and Jiaqi Wang.
\newblock {Long-CLIP}: Unlocking the long-text capability of {CLIP}.
\newblock In \emph{European Conference on Computer Vision (ECCV)}, volume 15109
  of \emph{Lecture Notes in Computer Science}, pp.\  310--325, Cham, 2024.
  Springer Nature Switzerland.
\newblock ISBN 978-3-031-72982-9.
\newblock \doi{10.1007/978-3-031-72983-6_18}.

\bibitem[Zhang et~al.(2025{\natexlab{a}})Zhang, Zhang, Wu, Wang, Wetzstein,
  Lin, and Liu]{threedgenbench}
Yuhan Zhang, Mengchen Zhang, Tong Wu, Tengfei Wang, Gordon Wetzstein, Dahua
  Lin, and Ziwei Liu.
\newblock {3DGen-Bench}: Comprehensive benchmark suite for {3D} generative
  models.
\newblock \emph{arXiv preprint arXiv:2503.21745}, 2025{\natexlab{a}}.

\bibitem[Zhang et~al.(2025{\natexlab{b}})Zhang, Cui, Yang, Li, and
  Xu]{hyperscore}
Yujie Zhang, Bingyang Cui, Qi~Yang, Zhu Li, and Yiling Xu.
\newblock Benchmarking and learning multi-dimensional quality evaluator for
  text-to-3d generation.
\newblock In \emph{IEEE/CVF International Conference on Computer Vision
  (ICCV)}, pp.\  18563--18574, 2025{\natexlab{b}}.
\newblock \doi{10.1109/ICCV51701.2025.01725}.

\bibitem[Zhao et~al.(2024)Zhao, Andrews, Papakyriakopoulos, and
  Xiang]{measuredatasetdiversity}
Dora Zhao, Jerone T.~A. Andrews, Orestis Papakyriakopoulos, and Alice Xiang.
\newblock Position: Measure dataset diversity, don't just claim it.
\newblock In \emph{International Conference on Machine Learning (ICML)}, volume
  235 of \emph{Proceedings of Machine Learning Research}, pp.\  60644--60673.
  PMLR, 2024.

\bibitem[Zhao et~al.(2026)Zhao, Jia, Je, Tang, Chan, Spedden, Palleschi, Huang,
  Wang, and Zheng]{defectbench}
Zhenyu Zhao, Nanshan Jia, Jihyeon Je, Yifu Tang, Alvin Chan, Michael Spedden,
  Michael~V. Palleschi, Sui Huang, Jingshen Wang, and Zeyu Zheng.
\newblock {3D-DefectBench}: A controlled factorial study of vision-language
  model evaluation pipelines for fine-grained 3d generation defects.
\newblock \emph{arXiv preprint arXiv:2607.10826}, 2026.

\end{thebibliography}

\appendix
\section{Appendix map}
\label{app:map}

This appendix keeps the main text focused on the audit logic and decision-level evidence.
Appendix~\ref{app:closest-work} separates our audit from the closest adjacent work.
Appendix~\ref{app:card-scope} states the configuration-card fields and scope limits.
Appendices~\ref{app:renders}--\ref{app:paraphrase-audit} provide render examples, caption grammar details, lighting scalars, protocol envelopes, selection-aware inference, hosted-score provenance, and paraphrase audit details.
Appendix~\ref{app:per-evaluator} gives the per-evaluator tables, and Appendix~\ref{app:bestlevel} the best and worst level per render factor.
Appendices~\ref{app:followup-factorial}--\ref{app:followup-power} report the preregistered caption-by-pitch factorial, the view-aggregation comparison, the full zero-cell re-scoring with the paraphrase-subset sensitivity, and the power analysis behind the confirmatory plan.
Appendices~\ref{app:case-study} onward add the observed protocol cards, the out-of-sample prompt replication, and the blinded human validation study.
Appendix~\ref{app:case-study} fills observed cards for two published comparisons and re-executes one published protocol inside the audit's configuration space.

\section{Closest-work boundary}
\label{app:closest-work}

Table~\ref{tab:closest-work} separates a new evaluator from an audit of the measurement pipeline.
``Not applicable'' marks a category outside the cited work's scope rather than a claimed gap in the full literature.

\begin{table*}[ht]
    \caption{Closest-work boundary.
        The cited work improves or meta-evaluates text, 2D, and 3D metrics, while the table distinguishes adjacent components from our combined intervention.
        ``Related'' marks an adjacent design rather than the same intervention.}
    \label{tab:closest-work}
    \centering
    \scriptsize
    \setlength{\tabcolsep}{2.5pt}
    \begin{tabular}{@{}p{0.16\textwidth}p{0.13\textwidth}p{0.13\textwidth}p{0.13\textwidth}p{0.13\textwidth}p{0.13\textwidth}p{0.10\textwidth}@{}}
        \toprule
        \textbf{Work}                                                           & \textbf{Artifact held fixed} & \textbf{Render factors isolated} & \textbf{Caption wording isolated} & \textbf{Decision stability} & \textbf{Controlled degradation} & \textbf{Reporting artifact} \\
        \midrule
        AutoMetrics~\citep{autometrics}                                         &
        Equivalent-quality outputs, not one frozen artifact                     &
        Not applicable to 3D rendering                                          &
        Output rephrasing, not a 3D caption wrapper                             &
        Metric validity, not generator-winner stability                         &
        LLM-generated worse-quality outputs                                     &
        Metric Cards, not a 3D configuration card                                                                                                                                                                                                                                   \\
        \addlinespace
        T2IScoreScore~\citep{t2iscorescore}                                     &
        Related 2D image families under a fixed prompt                          &
        Not applicable to 3D rendering                                          &
        Semantic error graphs, not caption wrappers                             &
        Metric ordering, not generator-winner stability                         &
        Semantic error graphs                                                   &
        No analogous card in the stated contribution                                                                                                                                                                                                                                \\
        \addlinespace
        3D-DefectBench~\citep{defectbench}                                      &
        Fixed assets across 84 inference designs                                &
        Camera protocol and visual input crossed                                &
        Prompt schema crossed                                                   &
        Factor effects and judge comparisons for defect detection               &
        Nine human-labeled defect categories                                    &
        Released data and metadata, not a reporting card                                                                                                                                                                                                                            \\
        \addlinespace
        Cross-Model VLM-Judge~\citep{asaria2026crossmodel}                      &
        Fixed mesh pairs across 24 rendered views                               &
        Fixed multi-view rig, not render-factor isolation                       &
        Fixed judge prompt, not a caption-wording audit                         &
        Pairwise asset preference, not generator-winner stability               &
        Face-drop degradation, not room-layout scrambling                       &
        Reproducible judge protocol, not a configuration card                                                                                                                                                                                                                       \\
        \addlinespace
        SceneCritic~\citep{scenecritic}                                         &
        Yes across views and repeated VLM calls                                 &
        Viewpoint and repeated calls, not eight OFAT factors                   &
        Prompt sensitivity motivated, not experimentally swept                  &
        View-dependent method-ranking reversals for one VLM judge               &
        No fixed-artifact degradation audit                                     &
        SceneOnto and a symbolic critic, not a configuration card                                                                                                                                                                                                                   \\
        \addlinespace
        T3Bench and Eval3D~\citep{t3bench,eval3d}                               &
        Generated assets, not a same-artifact audit                             &
        Many views, not a render-factor audit                                   &
        Caption or QA pipelines, not a wording audit                            &
        Leaderboards, not stability under configurations                        &
        Janus or inconsistency probes, not room-layout scrambling               &
        No analogous card in the stated contribution                                                                                                                                                                                                                                \\
        \addlinespace
        VIPScene / FPVScore~\citep{fpvscene}                                    &
        Generated scenes under fixed view rigs                                  &
        Top-down versus first-person, not an OFAT protocol-sensitivity audit    &
        Prompt-design ablation, not a caption-grammar sweep                     &
        Repeated-query and inter-model ranking agreement                        &
        System ablations, not paired layout edits                               &
        No analogous card in the stated contribution                                                                                                                                                                                                                                \\
        \addlinespace
        SceneEval and LEGO-Eval~\citep{sceneeval,legoeval}                      &
        Instruction-scene pairs, not a same-artifact configuration audit        &
        Render or tool evidence, not a factor audit                             &
        Structured constraints, not a wording audit                             &
        Method results, not stability under configurations                      &
        Misalignment or constraint tests, not controlled pairs                  &
        No analogous card in the stated contribution                                                                                                                                                                                                                                \\
        \specialrule{1.0pt}{0.35em}{0.35em}
        \textbf{This audit}                                                     &
        \textbf{Yes: \NumScenes{} frozen scenes}                                &
        \textbf{Resolution, focal length, lighting, background, pitch, and yaw} &
        \textbf{Template and paraphrase}                                        &
        \textbf{Ranks and top-1 under every factor}                             &
        \textbf{Object removal and layout scrambling}                           &
        \textbf{Evaluator Configuration Card}                                                                                                                                                                                                                                       \\
        \bottomrule
    \end{tabular}
\end{table*}

\subsection{Broader related work}
\label{app:broader-related}

Procedural scene generation includes Holodeck, Holodeck~2.0, HSM, and GLTreeSearch~\citep{holodeck,holodeck2,hsm,gltreesearch}.
Diffusion and instruction-driven alternatives include DiffuScene, InstructScene, FreeScene, and ArtiScene~\citep{diffuscene,instructscene,freescene,artiscene}.
Constraint programs, VLM optimization, and reasoning-driven construction broaden the generator space~\citep{constraintir,layoutvlm,idesign,sceneweaver,reason3d}.
Their evaluation choices also vary: top-down CLIPScore, perspective or isometric renders, MLLM judges, and symbolic relation checks appear across these systems.

Image-text audits have documented failures in attribute binding, unusual viewpoints, prompt wording, and compositional relations~\citep{sugarcrepe,colorswap,viewsnotequal,t2icompbench}.
Question-answering and unified VLM metrics seek more targeted evidence~\citep{dsg,t2icompbench,gecko,t2iscorescore}.
Object and mesh benchmarks use different units of comparison, including generated assets, pairwise asset preferences, or geometry defects~\citep{hyperscore,eval3d,gpteval3d,threedgenbench,db3dme}.
Scene-oriented benchmarks add relational, geometric, first-person, and trajectory evidence~\citep{relscene,legoeval,fpvscene,worldscore}.
These studies motivate broader validation, while our main-text comparison is restricted to protocol effects on fixed structured room scenes.

\section{Evaluator Configuration Card and scope details}
\label{app:card-scope}

Table~\ref{tab:evaluator-card} turns the reporting-artifact column of Table~\ref{tab:closest-work} into the fields needed to audit a single-image text-to-3D evaluator report.
This domain-specific run-side contract complements BenchmarkCards, Eval Factsheets, and Evaluation Cards rather than claiming the broader card concept~\citep{sokol2025benchmarkcards,bordes2025evalfactsheets,evalcards}.
\begin{table}[ht]
    \caption{Paper-ready Evaluator Configuration Card checklist.
        A complete JSON card records these fields, validates against schema version \texttt{1.0}, and receives a deterministic \texttt{ecc-v1-sha256:\ldots} identifier.}
    \label{tab:evaluator-card}
    \centering
    \footnotesize
    \setlength{\tabcolsep}{3pt}
    \begin{tabular}{@{}p{0.27\linewidth}p{0.65\linewidth}@{}}
        \toprule
        \textbf{Card field}          & \textbf{Required report content}                                                                                                   \\
        \midrule
        Schema and status            &
        \texttt{card\_type}, \texttt{schema\_version}, evidence status, and the validator-produced card identifier.                                                       \\
        Scene artifact source        &
        Dataset, generator set, scene and prompt counts, frozen-scene rule, and artifact provenance.                                                                      \\
        Render protocol              &
        Renderer, camera pose, focal length, resolution, lighting, background, random seeds, and post-processing.                                                         \\
        Caption protocol             &
        Prompt template, paraphrase rule, prompt source, and whether wording was frozen before scoring.                                                                   \\
        View policy                  &
        Fixed, averaged, or selected view, declared views, and selection rule.                                                                                            \\
        Evaluator identity           &
        Model or metric name, checkpoint/API snapshot, instruction, decoding settings, and call date.                                                                     \\
        Score uncertainty            &
        Interval or bootstrap rule, resampling unit, comparison family, decision thresholds, multiplicity policy, and whether claims cover scores, ranks, or top methods. \\
        Rank stability               &
        Rank range or winner changes across declared configurations.                                                                                                      \\
        Companion check              &
        Geometry-grounded, multi-view, human, or explicit \texttt{not\_reported} companion validation.                                                                    \\
        Released or missing evidence &
        Frozen scenes, render grids, raw score files, evaluator responses, hash manifest, and any known missing evidence.                                                 \\
        \bottomrule
    \end{tabular}
\end{table}

The audit is intrinsic and does not estimate which score difference a person would consider meaningful.
Its configuration sweeps keep the scene and task prompt fixed, but caption wrappers and render settings can change the evaluator's input semantics or visible evidence.
The OFAT design does not estimate interactions, and external validity requires blinded human comparison of representative scenes, which Appendix~\ref{app:human-study} supplies in part for intact, scrambled, and camera-varied scenes.
The judge results use one GPT-4o configuration and one rating instruction, so they do not isolate model capability or establish generalization to other MLLMs.
The rank and $\rho$ results are properties of this six-generator field and an $N = 50$ sample, while every effect magnitude remains specific to the Blender pipeline and prompt sample.

\section{Qualitative renders}
\label{app:renders}

\begin{figure}[ht]
    \centering
    \includegraphics[width=\linewidth]{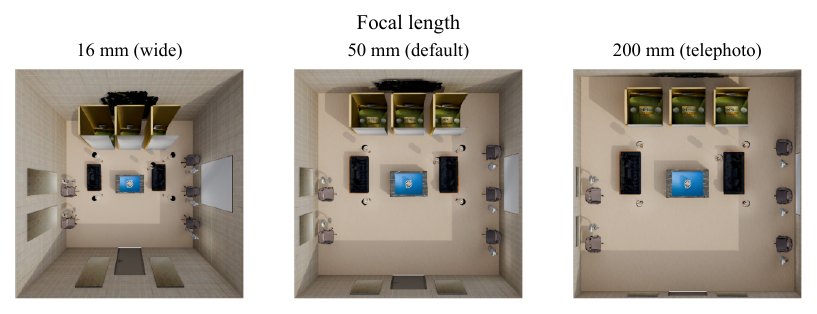}\\
    \includegraphics[width=\linewidth]{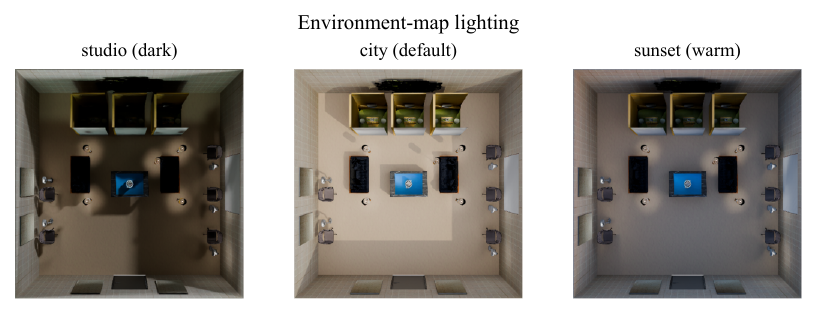}
    \caption{The same scene under two configuration factors.
        Top: focal length $16$/$50$/$200$\,mm at a fixed, auto-framed top-down view.
        Bottom: environment-map lighting (studio/city/sunset).
        Figure~\ref{fig:thesis} shows the pitch sweep in the main paper.}
    \label{fig:montage-render}
\end{figure}

Figure~\ref{fig:montage-render} shows the same generated scene under two render-side configuration factors, making concrete that these are ordinary choices rather than exotic manipulations.

\section{Template grammar}
\label{app:templates}

\begin{figure}[ht]
    \centering
    \includegraphics[width=\linewidth]{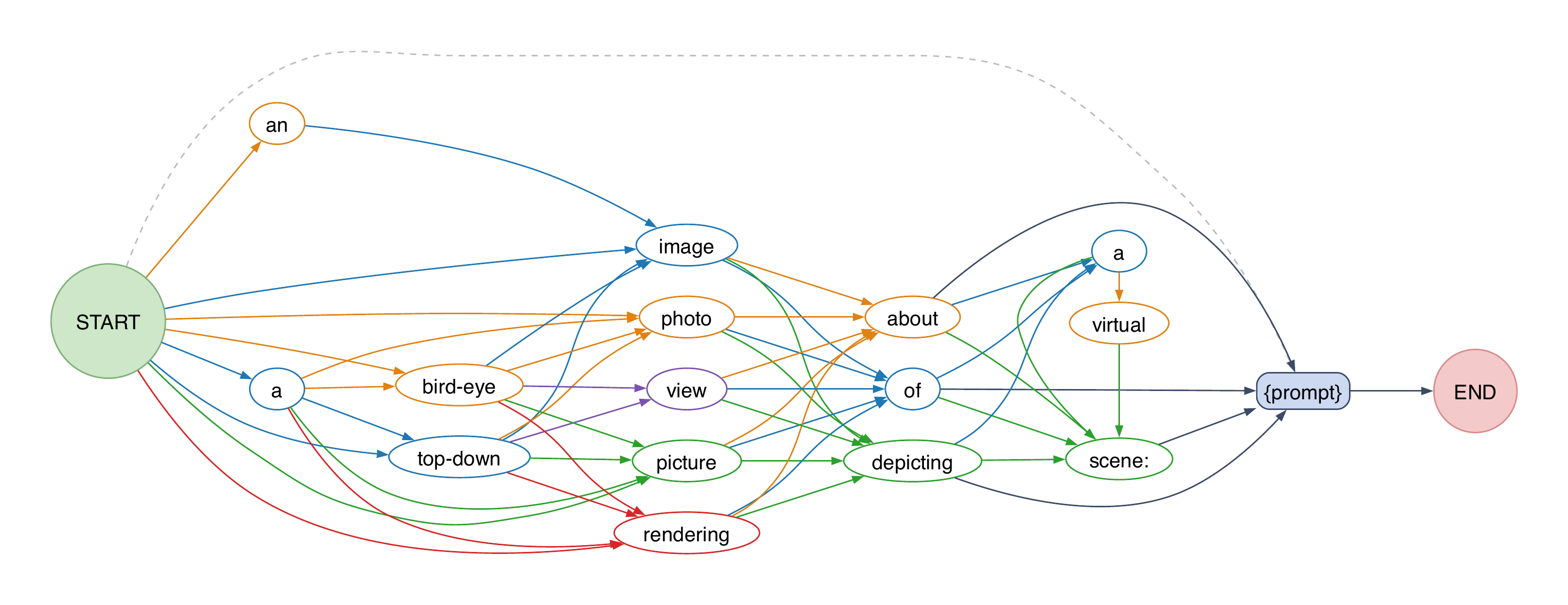}
    \caption{The template grammar as a directed acyclic token graph.
        Each of the $337$ templates is exactly one source-to-sink path (START\,$\rightarrow \cdots \rightarrow$\,\{prompt\}\,$\rightarrow$\,END).
        Each edge is colored by the node it points into, and every node is outlined in its own color, so an edge and its target always match.
        The dashed edge from START to \{prompt\} is the degenerate identity template.}
    \label{fig:template-dag}
\end{figure}

The 337 evaluation templates of Section~\ref{sec:method-text} are the source-to-sink paths of a small grammar.
Table~\ref{tab:slotgrammar} reads the grammar as an ordered sequence of slots (the form used to build a caption by hand), and Figure~\ref{fig:template-dag} draws the same grammar as a directed acyclic token graph (the form that makes the path count explicit).
The token nodes are written in lower case, but the first letter of every realized caption is capitalized, so the default template reads \emph{``A top-down view of \{prompt\}''} rather than \emph{``a top-down\,\ldots''}.
The one exception is the degenerate identity path, whose only token is the prompt, which is emitted verbatim.

\begin{table}[ht]
    \caption{The template grammar as ordered slots.
        A caption picks at most one entry from each slot in order (subject to the adjacency in Figure~\ref{fig:template-dag}) and always ends with the prompt.
        The connective is present exactly when a head noun is.
        The default template \emph{``a top-down view of \{prompt\}''} takes the article, the \emph{top-down} viewpoint, \emph{view}, the connective \emph{of}, no scene frame, and the prompt.}
    \label{tab:slotgrammar}
    \centering
    \small
    \begin{tabular}{@{}lll@{}}
        \toprule
        \textbf{Slot} & \textbf{Status} & \textbf{Choices} \\
        \midrule
        Article & \textit{optional} & a, an \\
Viewpoint & \textit{optional} & top-down, bird-eye \\
Head noun & \textit{optional} & image, photo, picture, rendering, view \\
Connective & \textit{with a noun} & of, about, depicting \\
Scene frame & \textit{optional} & scene:, a scene:, a virtual scene: \\
Prompt & \textit{required} & \{prompt\}
                         \\
        \bottomrule
    \end{tabular}
\end{table}

\subsection{Which wording the evaluators reward}
\label{app:template-effect}

Section~\ref{sec:results-text} reports how far the template moves the score.
Here we ask which choices in the grammar are responsible, by parsing all \NumTemplates{} captions back into their slots and averaging the score of each slot choice over the text-swept evaluators.
Table~\ref{tab:slotgrammar-effect} gives, per slot, its dispersion and each choice's mean shift from the slot's own average.
To keep a slot from being confounded by the company it keeps, each is measured on a matched subset.
Head nouns are compared only where a viewpoint word is present, so all five nouns span the same caption skeletons, and connectives only where a head noun is present.
The head-noun choice is not neutral.
\emph{Rendering} scores highest and \emph{photo} lowest, a range of \TmplNounDisp\% of the slot mean, and rendering leads for all \NounSweepN{} evaluators.
Because these images are renders, part of this gap could reflect appropriate image-type matching rather than an evaluator defect.
It still changes the reported scale without changing the artifact, so matching the caption convention is necessary for a numerical comparison.
The scene frame matters at least as much: naming \emph{a virtual scene:} sits about $8\%$ above the slot average and omitting it about $4\%$ below, a range of \TmplFrameDisp\%.
Naming a viewpoint at all is worth more than which viewpoint, since \emph{top-down} beats \emph{bird-eye} by only a few percent while dropping the viewpoint word costs about $17\%$ relative to \emph{top-down}, nearly the slot's full range.
The article and the connective are close to inert.
The article deserves a note, because its raw marginal is misleading.
\emph{An} occurs only before the vowel-initial \emph{image} and never beside a viewpoint word, so every \emph{An image} caption also forgoes the viewpoint, and lumping them together makes \emph{an} look like a large penalty.
Measured among viewpoint-free captions, where \emph{a}, \emph{an}, and no article are genuine alternatives, all three fall within a few percent, and the choice between \emph{a} and \emph{an} is in any case fixed by the following word rather than free.
These choices compound.
The best template varies in its exact wording, but 16 of 19 evaluators put a \emph{top-down} template first, every evaluator's best names a head noun, and 18 of 19 use the frame \emph{a virtual scene:}.
The per-evaluator best beats the default \emph{``a top-down view of''} by a median of \BestTmplInfl\% (up to \BestTmplInflMax\%), with the default ranking only \DefaultTmplRank{} of \NumTemplates.
An author who reports the same renders under the best caption rather than the default raises every alignment number without touching a pixel, which is the template analogue of the best-view inflation of Section~\ref{sec:results-bestview}.

\section{Environment-map scalars}
\label{app:envmaps}

Table~\ref{tab:envmaps} lists the two scalars used to summarize the eight lighting maps in Section~\ref{sec:results-render}: a log-average exposure value (EV, computed from the map's luminance) and a correlated color temperature (CCT, from the chromaticity of the map's total radiated energy).
The rightmost column is each map's mean score shift, standardized per evaluator and averaged over evaluators and scenes, so that maps with a positive value tend to raise the score and maps with a negative value tend to depress it.
Brightness (EV) correlates $r = \EnvEvCorr$ with this aggregate shift, but this association does not isolate brightness from other map properties or establish the same effect for every scene.

\begin{table}[ht]
    \caption{The eight lighting maps ordered dark to bright.
        EV is a log-average exposure value (lower is darker).
        CCT is correlated color temperature (lower is warmer).
        The default map ``city'' is the brightest.
        Mean $z$-shift is the map's average standardized score shift across evaluators and scenes.}
    \label{tab:envmaps}
    \centering
    \small
    \begin{tabular}{@{}lrrr@{}}
        \toprule
        \textbf{Environment map} & \textbf{EV} & \textbf{CCT (K)} & \textbf{Mean $z$-shift} \\
        \midrule
        studio & -6.41 & 7802 & -0.086 \\
night & -5.13 & 5051 & -0.212 \\
courtyard & -3.70 & 5759 & -0.002 \\
sunrise & -3.24 & 6004 & -0.043 \\
forest & -2.22 & 7513 & +0.093 \\
interior & -2.20 & 5420 & +0.081 \\
sunset & -2.02 & 8785 & +0.024 \\
city$^\star$ & -1.17 & 6444 & +0.145 \\
                                                              \\[-1.2em]
        \bottomrule
    \end{tabular}
\end{table}

\section{Pairwise comparison over the tested protocols}
\label{app:protocol-margins}

Table~\ref{tab:protocol-margins} applies the protocol margin envelope to every unordered generator pair.
For an alignment evaluator, the audit set contains the baseline and all 386 shared levels of the seven rankable factors, with yaw measured at the documented $45^\circ$ pitch.
Repeated default-equivalent levels are retained as recorded protocols and do not alter the envelope endpoints.
Scene-specific paraphrases are excluded because they are not a shared configuration level.
The quality control has no caption sweep, so its 50 recorded baseline and render protocols are reported separately.
The categories retain exact aggregate ties and describe the saved point estimates only.

\begin{table}[ht]
    \caption{Pairwise ordering over the finite tested protocol set.
        Strict ordering means that a generator pair keeps the same nonzero margin sign throughout the set.
        A sign crossing includes both positive and negative margins, while a tie boundary reaches zero without crossing signs.
        A stable winner beats all five alternatives at every tested protocol.
        These descriptive ranges are not confidence intervals or statements of human-grounded superiority.
        The hosted judge's zero-score handling is documented in Appendix~\ref{app:hosted-provenance}, and the image-only control, marked \emph{qc}, lacks a caption sweep and is summarized over its 50 render-only protocols.}
    \label{tab:protocol-margins}
    \centering
    \small
    \setlength{\tabcolsep}{4pt}
    \begin{tabular}{@{}lrrrrl@{}}
        \toprule
        \textbf{Evaluator} & \textbf{Protocols} & \textbf{Strict ordering} & \textbf{Sign crossing} & \textbf{Tie boundary} & \textbf{Stable winner} \\
        \midrule
CLIP ViT-B/32 & 387 & 2/15 & 13/15 & 0/15 & none \\
CLIP ViT-B/16 & 387 & 0/15 & 15/15 & 0/15 & none \\
CLIP ViT-L/14 & 387 & 0/15 & 15/15 & 0/15 & none \\
CLIP ViT-L/14@336 & 387 & 1/15 & 14/15 & 0/15 & none \\
OpenCLIP ViT-L/14 & 387 & 0/15 & 15/15 & 0/15 & none \\
OpenCLIP ViT-H/14 & 387 & 0/15 & 15/15 & 0/15 & none \\
OpenCLIP ViT-bigG/14 & 387 & 1/15 & 14/15 & 0/15 & none \\
MetaCLIP ViT-L/14 & 387 & 4/15 & 11/15 & 0/15 & none \\
DFN ViT-L/14 & 387 & 7/15 & 8/15 & 0/15 & none \\
EVA-CLIP ViT-L/14 & 387 & 2/15 & 13/15 & 0/15 & none \\
CLIPA ViT-L/14 & 387 & 7/15 & 8/15 & 0/15 & none \\
Long-CLIP ViT-L/14 & 387 & 1/15 & 14/15 & 0/15 & none \\
SigLIP ViT-L/16 & 387 & 3/15 & 12/15 & 0/15 & none \\
SigLIP 2 ViT-L/16 & 387 & 5/15 & 10/15 & 0/15 & none \\
BLIP-2 ITM & 387 & 1/15 & 14/15 & 0/15 & none \\
BLIP-2 ITC & 387 & 3/15 & 12/15 & 0/15 & none \\
BLIP-VQA & 387 & 1/15 & 14/15 & 0/15 & none \\
VQAScore & 387 & 3/15 & 12/15 & 0/15 & none \\
GPT-4o judge & 387 & 7/15 & 8/15 & 0/15 & Scenethesis \\
CLIP-IQA & 50 & 0/15 & 15/15 & 0/15 & none\textsuperscript{qc} \\
                                                                                                              \\[-1.2em]
        \bottomrule
    \end{tabular}
\end{table}

\section{Selection-aware same-pair inference}
\label{app:selection-inference}

A changed sample winner need not imply a well-separated reversal, especially after searching many configurations.
We construct a single contrast family containing all 19 alignment evaluators, 15 generator pairs, and 387 configurations: the baseline plus the 386 shared levels of the seven rankable factors.
This gives $J=\AuditNContrasts$ paired contrasts, excluding the image-only control and the prompt-specific paraphrase strings.
For contrast $j$, let $d_{p,j}$ be the score difference between its two generators on prompt $p$, with empirical mean $\bar d_j$ and paired standard error $\widehat{\mathrm{SE}}_j$.
We resample the 50 prompt blocks jointly across every evaluator, method, and configuration, preserving all within-prompt dependencies.
Using $B=\AuditNBootstrap$ draws with seed 42, we compute the bootstrap maximum standardized deviation and its empirical $95\%$ quantile:
\begin{equation}
    T_b = \max_{j:\widehat{\mathrm{SE}}_j>0} \frac{|\bar d_j^{*(b)}-\bar d_j|}{\widehat{\mathrm{SE}}_j}, \qquad I_j = \left[\bar d_j-q_{0.95}\widehat{\mathrm{SE}}_j,\ \bar d_j+q_{0.95}\widehat{\mathrm{SE}}_j\right].
\end{equation}
The standard errors are fixed at their original-sample values, not re-estimated within each draw, and exact-constant contrasts are ineligible for inferential declarations.
The resulting critical value is $q_{0.95}=\AuditMaxCritical$.
This is approximate exploratory bootstrap coverage for the fixed six-generator field and empirical prompt distribution, not a finite-sample family-wise guarantee or a prevalence estimate for all scene generators.

For each evaluator and factor, we consider pairs that are distinct unique sample winners at two levels and ask whether their two intervals exclude zero in opposite directions.
All candidate pairs and levels are already in the simultaneous family, including those subsequently selected for display.
The primary seed yields \AuditPointEstimateReversals{} evaluators with descriptive reversals, \AuditPointwiseReversals{} with opposite pointwise intervals, and \AuditSimultaneousReversals{} with opposite simultaneous intervals (Table~\ref{tab:reversal-examples}).
At seed 1234, the pointwise count is eight because the BLIP-VQA example is marginal, while the descriptive and simultaneous counts are unchanged.
The baseline-only comparison in the primary analysis yields four evaluators with opposite pointwise intervals, a different estimand from searching two arbitrary levels of the same factor.
None of these pairwise comparisons establishes that either selected winner is superior to all five alternatives.

\section{Hosted-score provenance and zero-score sensitivity}
\label{app:hosted-provenance}

The GPT-4o score file records 120,270 stage-score cells, of which 2,333 are zero.
These zero cells are dependent measurements, not independent model runs.
The scoring client returns zero for request or parsing exceptions, so a stored zero can represent a genuine zero rating or a failed request.
We therefore re-score the zero cells among the 300 audited scramble pairs and report deterministic sensitivity bounds that cover both interpretations.
Table~\ref{tab:gpt-outcomes} decomposes the 300 paired scramble outcomes instead of equating ties with detected degradation.

\begin{table}[ht]
    \caption{GPT-4o scramble outcomes from the same 300 intact and scrambled score pairs.
        Intervals use the joint prompt bootstrap of Appendix~\ref{app:selection-inference} and are pointwise descriptive summaries.
        Tie-adjusted discrimination credits a drop as one, a tie as one-half, and a rise as zero.}
    \label{tab:gpt-outcomes}
    \centering
    \small
    \begin{tabular}{@{}lrr@{}}
        \toprule
        \textbf{Outcome}            & \textbf{Count} & \textbf{Percent [95\% interval]}                            \\
        \midrule
        Strict drop                 & \AuditGptDrops & \AuditGptDropPct{} [\AuditGptDropLo, \AuditGptDropHi]       \\
        Exact tie                   & \AuditGptTies  & \AuditGptTiePct{} [\AuditGptTieLo, \AuditGptTieHi]          \\
        Rise                        & \AuditGptRises & \AuditGptRisePct{} [\AuditGptRiseLo, \AuditGptRiseHi]       \\
        Tie-adjusted discrimination &                & \AuditGptCreditPct{} [\AuditGptCreditLo, \AuditGptCreditHi] \\
        \bottomrule
    \end{tabular}
\end{table}

There are 4 zero baselines, 9 zero scrambled scores, and 2 pairs with both scores zero, giving \AuditGptZeroPairs{} pairs involving at least one zero.
The remaining \AuditGptNonzeroPairs{} pairs contribute 161 units of directional credit, yielding $55.7\%$ on this selected subset.
That subset is not an unbiased failure-corrected estimate.
Allowing arbitrary credit for every zero-involved pair instead bounds full-sample discrimination by $[161/300,172/300]=[\AuditGptSensitivityLo,\AuditGptSensitivityHi]\%$.
These bounds capture uncertainty about the zero scores, not sampling uncertainty.

We re-score the 13 stored zero-valued cells, 4 intact-side and 9 scrambled-side, with GPT-4o at temperature 0 under a bounded retry policy that treats a zero-valued response as a suspected call failure and repeats it for up to six attempts.
Eight cells return a nonzero score within the retry limit with a mean resolved score of 15.6, and the remaining five stay zero across every attempt.
Every persistent zero carries an explicit low-alignment justification in its reasoning, naming the requested objects missing from the render, and the five persist in one HSM, one LayoutVLM, and three Reason-3D scenes.
The retry policy resolves the unstable zeros, while the persistent ones are consistent with genuine zero ratings under this re-scoring run.
The deterministic sensitivity range of Table~\ref{tab:gpt-outcomes} bounds the effect of the zero-involved pairs on the full-sample estimand.

\section{Paraphrase audit}
\label{app:paraphrase-audit}

The frozen prompt file stores 10 rewrites for each of the 50 prompts, and we audit every rewrite with two hosted auditors at temperature 0.
The auditors are \texttt{deepseek-v4-flash}, the model that generated the rewrites, and the independent hosted judge GPT-4o.
Each auditor receives the original prompt and one rewrite and answers a fixed rubric question: does the rewrite preserve every spatial and semantic requirement of the original, including objects, counts, spatial relations, room type, and style.
The auditors accept 245 of 500 rewrites ($49.0\%$) and 238 of 500 ($47.6\%$), and they agree on 376 of 500 rewrites ($75.2\%$).
The similarity score stored with each rewrite does not separate the two outcomes: mean similarity is 0.236 for accepted and 0.237 for rejected rewrites under one auditor, and 0.232 versus 0.240 under the other.
A typical joint rejection rewrites \emph{``a bar''} as \emph{``a pub''} or \emph{``a tavern''}, which changes the requested object.
A machine verdict is not a human judgment.

\section{Per-evaluator detail}
\label{app:per-evaluator}

\begin{table}[ht]
    \caption{Evaluator-configuration factors, each swept independently with all others held at the default $c_0$ ($\star$).
        The cross-factor baseline uses a white background.
        The background sweep ranges over the listed gray and chromatic fills.}
    \label{tab:factors}
    \centering
    \small
    \begin{tabular}{@{}llp{8.1cm}@{}}
        \toprule
        \textbf{Factor}                    & \textbf{\#} & \textbf{Levels}                                                                              \\
        \midrule
        Resolution (pixel)                 & 9           & 196, 224, 256, 336, 384, 448, 512$^\star$, 768, 1024                                         \\
        Focal length (mm)                  & 7           & 16, 24, 35, 50$^\star$, 85, 100, 200                                                         \\
        Lighting (environment map)         & 8           & city$^\star$, courtyard, forest, interior, night, studio, sunrise, sunset                    \\
        Background (RGB)                   & 10          & 0, 65, 118, 128, 186, 204, 255$^\star$ gray, plus red, green, blue                           \\
        Pitch (degrees)                    & 7           & 0$^\star$, 15, 30, 45, 60, 75, 90                                                            \\
        Yaw (degrees, at pitch $45^\circ$) & 8           & 0$^\star$, 45, 90, 135, 180, 225, 270, 315                                                   \\
        Template                           & 337         & source-to-sink paths of the grammar (Appendix~\ref{app:templates}).
        Defaults to \emph{``a top-down view of \{\,$p'$\,\}''}                                                                                          \\
        Paraphrase                         & 8--11       & distinct strings from the original prompt plus 10 intended semantics-preserving LLM rewrites \\
        \bottomrule
    \end{tabular}
\end{table}

\begin{table}[ht]
    \caption{The 20 evaluators under audit, grouped by family.
        CLIP-IQA is reported apart from the alignment metrics because it does not condition on the prompt.}
    \label{tab:evaluators}
    \centering
    \small
    \begin{tabular}{@{}lp{8.9cm}@{}}
        \toprule
        \textbf{Group}       & \textbf{Members}                                                                                                                                                       \\
        \midrule
        CLIP (OpenAI)        & ViT-B/32, ViT-B/16, ViT-L/14, ViT-L/14@336~\citep{clip}                                                                                                                \\
        CLIP (open data)     & OpenCLIP ViT-L/14, ViT-H/14, ViT-bigG/14~\citep{cherti2023reproducible}, MetaCLIP~\citep{metaclip}, DFN~\citep{dfn}, EVA-CLIP~\citep{evaclip}, and CLIPA~\citep{clipa} \\
        CLIP (long-text)     & Long-CLIP~\citep{longclip}                                                                                                                                             \\
        Sigmoid objective    & SigLIP~\citep{siglip} and SigLIP 2~\citep{siglip2}                                                                                                                     \\
        BLIP-2~\citep{blip2} & ITM and ITC heads                                                                                                                                                      \\
        VQA-based            & BLIP-VQA~\citep{t2icompbench} and VQAScore~\citep{vqascore}                                                                                                            \\
        MLLM judge           & GPT-4o                                                                                                                                                                 \\
        Quality (scoped)     & CLIP-IQA~\citep{clipiqa}                                                                                                                                               \\
        \bottomrule
    \end{tabular}
\end{table}

\begin{figure}[ht]
    \centering
    \includegraphics[width=0.95\linewidth]{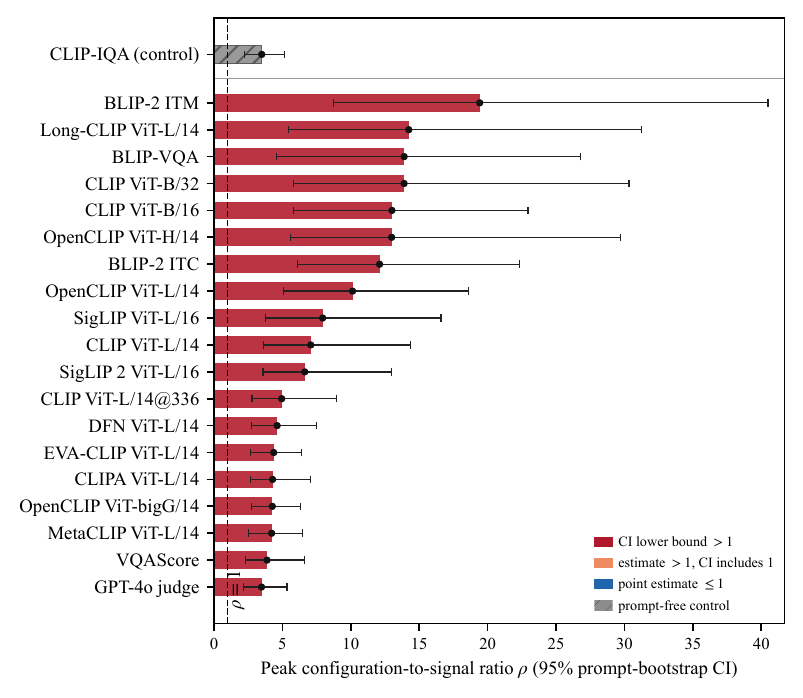}
    \caption{Peak configuration-to-signal ratio for the fixed six-generator field, with per-evaluator prompt-bootstrap $95\%$ intervals (50 prompts with generators fixed).
        The lower bound exceeds one for all 19 alignment evaluators.
        CLIP-IQA is shown separately as a prompt-free quality control.
        A high ratio alone does not establish a changed pairwise ordering.}
    \label{fig:rho-bars}
\end{figure}

Tables~\ref{tab:dispersion}, \ref{tab:rho}, \ref{tab:ranking}, and~\ref{tab:content} give the per-factor numbers summarized in Table~\ref{tab:headline}: full dispersion, the configuration-to-signal ratio $\rho$, rank stability, and tie-adjusted directional discrimination.
Figure~\ref{fig:rho-bars} visualizes prompt-bootstrap uncertainty for peak $\rho$.
Table~\ref{tab:rho-ci} also reports leave-one-generator-out sensitivity.

\paragraph{Additional result details.}
Within the OpenCLIP family, moving from ViT-L/14 to ViT-bigG/14 raises tie-adjusted scramble discrimination from \OpenClipLScr\% to \BigGScr\%, but also lowers template $\rho$ from \OpenClipLTmplRho{} to \BigGTmplRho.
The $\rho$ denominator uses the population variance over the six generator means.
Using sample variance would rescale every $\rho$ by $5/6$ and still leaves \NRhoAlignSampleGtOne{} alignment evaluators and the prompt-free control above $1$.
The GPT-4o scramble outcomes are decomposed into strict drops, exact ties, and rises in Appendix~\ref{app:hosted-provenance}.

\paragraph{Benchmark-source robustness.}
For the hosted GPT-4o judge, template sensitivity and weak scramble discrimination also occur outside the dominant SceneEval prompts.
Splitting that judge by prompt source, template dispersion is \GptTmplSE\% on SceneEval versus \GptTmplOther\% on the other three benchmarks, and scramble directional discrimination is \GptScrSE\% versus \GptScrOther\%.
Template sensitivity is stronger off SceneEval for this judge, and scramble discrimination is similarly weak on both source groups, but this source breakout does not establish the same pattern for all 20 evaluators.

\begin{table}[ht]
    \caption{Score dispersion (normalized range, \%) per evaluator and factor.
        Larger is worse.
        \emph{Res.} is resolution, \emph{Bg.} is background, \emph{Tmpl.} is template, and \emph{Para.} is paraphrase.}
    \label{tab:dispersion}
    \centering
    \small
    \setlength{\tabcolsep}{4pt}
    \begin{tabular}{@{}lcccccccc@{}}
        \toprule
        \textbf{Evaluator} & \textbf{Res.} & \textbf{Focal} & \textbf{Light} & \textbf{Bg.} & \textbf{Pitch} & \textbf{Yaw} & \textbf{Tmpl.} & \textbf{Para.} \\
        \midrule
        CLIP ViT-B/32 & 7 & 14 & 12 & 8 & 22 & 16 & 26 & 13 \\
CLIP ViT-B/16 & 9 & 12 & 12 & 6 & 21 & 15 & 30 & 12 \\
CLIP ViT-L/14 & 15 & 20 & 19 & 11 & 29 & 23 & 48 & 18 \\
CLIP ViT-L/14@336 & 13 & 16 & 16 & 10 & 27 & 19 & 48 & 17 \\
OpenCLIP ViT-L/14 & 19 & 18 & 19 & 10 & 38 & 25 & 52 & 16 \\
OpenCLIP ViT-H/14 & 15 & 15 & 18 & 12 & 37 & 24 & 57 & 18 \\
OpenCLIP ViT-bigG/14 & 10 & 10 & 12 & 10 & 26 & 17 & 39 & 12 \\
MetaCLIP ViT-L/14 & 14 & 17 & 19 & 11 & 33 & 27 & 47 & 14 \\
DFN ViT-L/14 & 15 & 15 & 15 & 11 & 39 & 24 & 52 & 15 \\
EVA-CLIP ViT-L/14 & 11 & 13 & 14 & 10 & 28 & 21 & 35 & 12 \\
CLIPA ViT-L/14 & 9 & 13 & 14 & 7 & 36 & 21 & 50 & 16 \\
Long-CLIP ViT-L/14 & 8 & 12 & 10 & 5 & 22 & 12 & 34 & 10 \\
SigLIP ViT-L/16 & 37 & 45 & 56 & 32 & 96 & 66 & 124 & 51 \\
SigLIP 2 ViT-L/16 & 21 & 24 & 23 & 14 & 65 & 46 & 91 & 31 \\
BLIP-2 ITM & 50 & 77 & 97 & 61 & 168 & 138 & 173 & 87 \\
BLIP-2 ITC & 8 & 14 & 15 & 8 & 29 & 17 & 55 & 17 \\
BLIP-VQA & 21 & 27 & 33 & 15 & 40 & 33 & 32 & 32 \\
VQAScore & 34 & 41 & 44 & 18 & 54 & 42 & 72 & 36 \\
GPT-4o judge & 62 & 66 & 63 & 53 & 101 & 82 & 97 & 63 \\
CLIP-IQA & 60 & 40 & 65 & 29 & 73 & 58 & -- & -- \\
                                                                                                                             \\[-1.2em]
        \bottomrule
    \end{tabular}
\end{table}

\begin{table}[ht]
    \caption{Configuration-to-signal ratio $\rho$ per evaluator and factor.
        Values $\ge 1$ mean that within-scene configuration variance equals or exceeds the variance between generator mean scores on that evaluator's scale.
        \emph{Res.} is resolution, \emph{Bg.} is background, \emph{Tmpl.} is template, and \emph{Para.} is paraphrase.}
    \label{tab:rho}
    \centering
    \small
    \setlength{\tabcolsep}{4pt}
    \begin{tabular}{@{}lcccccccc@{}}
        \toprule
        \textbf{Evaluator} & \textbf{Res.} & \textbf{Focal} & \textbf{Light} & \textbf{Bg.} & \textbf{Pitch} & \textbf{Yaw} & \textbf{Tmpl.} & \textbf{Para.} \\
        \midrule
        CLIP ViT-B/32 & 1.25 & 4.78 & 3.74 & 1.63 & 13.89 & 7.33 & 6.80 & 3.36 \\
CLIP ViT-B/16 & 1.59 & 3.93 & 3.65 & 0.95 & 13.01 & 5.98 & 9.35 & 2.84 \\
CLIP ViT-L/14 & 1.41 & 3.48 & 2.58 & 0.82 & 7.08 & 3.50 & 7.04 & 1.85 \\
CLIP ViT-L/14@336 & 0.85 & 1.58 & 1.38 & 0.47 & 4.69 & 1.83 & 4.95 & 1.28 \\
OpenCLIP ViT-L/14 & 1.80 & 2.15 & 2.02 & 0.63 & 10.13 & 3.31 & 6.53 & 1.24 \\
OpenCLIP ViT-H/14 & 1.70 & 2.11 & 2.73 & 1.37 & 12.99 & 3.98 & 12.96 & 2.28 \\
OpenCLIP ViT-bigG/14 & 0.42 & 0.54 & 0.65 & 0.44 & 4.26 & 1.18 & 3.08 & 0.59 \\
MetaCLIP ViT-L/14 & 0.64 & 1.22 & 1.31 & 0.48 & 4.22 & 2.42 & 3.48 & 0.60 \\
DFN ViT-L/14 & 0.55 & 0.70 & 0.60 & 0.30 & 4.62 & 1.27 & 3.24 & 0.50 \\
EVA-CLIP ViT-L/14 & 0.64 & 1.13 & 1.20 & 0.52 & 4.37 & 2.11 & 3.11 & 0.71 \\
CLIPA ViT-L/14 & 0.26 & 0.63 & 0.58 & 0.13 & 4.29 & 1.11 & 3.51 & 0.66 \\
Long-CLIP ViT-L/14 & 1.43 & 4.32 & 2.96 & 0.62 & 14.24 & 4.05 & 12.43 & 2.67 \\
SigLIP ViT-L/16 & 0.92 & 1.90 & 2.23 & 0.71 & 7.94 & 2.71 & 6.89 & 2.02 \\
SigLIP 2 ViT-L/16 & 0.71 & 1.09 & 0.94 & 0.36 & 6.64 & 2.37 & 6.55 & 1.33 \\
BLIP-2 ITM & 1.05 & 3.33 & 3.54 & 1.15 & 12.72 & 5.16 & 19.42 & 3.51 \\
BLIP-2 ITC & 0.50 & 1.61 & 1.68 & 0.43 & 7.69 & 2.09 & 12.11 & 2.21 \\
BLIP-VQA & 2.44 & 4.68 & 5.17 & 1.25 & 13.90 & 9.28 & 5.26 & 8.42 \\
VQAScore & 1.09 & 1.72 & 1.65 & 0.28 & 3.88 & 2.14 & 1.64 & 1.09 \\
GPT-4o judge & 1.14 & 1.41 & 1.07 & 0.68 & 3.48 & 2.05 & 0.87 & 0.99 \\
CLIP-IQA & 3.48 & 1.61 & 2.71 & 0.94 & 3.49 & 2.26 & -- & -- \\
                                                                                                                                    \\[-1.2em]
        \bottomrule
    \end{tabular}
\end{table}

\begin{table}[ht]
    \caption{Uncertainty and generator-set sensitivity of the configuration-to-signal ratio, and baseline generator separability.
        \emph{Peak $\rho$} is the point estimate of Table~\ref{tab:headline}.
        \emph{Prompt 95\% CI} resamples the $N = 50$ prompts with the six generators fixed.
        \emph{Leave-one-gen range} recomputes peak $\rho$ after dropping each generator.
        \emph{Top-1/2 gap} is the paired-bootstrap difference between the two baseline leaders, and $^\ast$ marks an interval that excludes zero.
        This column does not identify the actual winner pair at every configuration level.
        CLIP-IQA has no alignment Top-1/2 gap because it never conditions on the prompt, and its other rank columns are quality-control pseudo-rankings.}
    \label{tab:rho-ci}
    \centering
    \small
    \setlength{\tabcolsep}{4pt}
    \begin{tabular}{@{}lrccl@{}}
        \toprule
        \textbf{Evaluator} & \textbf{Peak $\rho$} & \textbf{Prompt 95\% CI} & \textbf{Leave-one-gen range} & \textbf{Top-1/2 gap [95\% CI]} \\
        \midrule
        CLIP ViT-B/32 & 13.89 & [5.81, 30.34] & [9.06, 49.20] & +0.91 [+0.12, +1.73]$^\ast$ \\
CLIP ViT-B/16 & 13.01 & [5.82, 22.94] & [7.92, 22.73] & +0.33 [-0.64, +1.28] \\
CLIP ViT-L/14 & 7.08 & [3.60, 14.39] & [5.98, 9.60] & +0.07 [-1.04, +1.05] \\
CLIP ViT-L/14@336 & 4.95 & [2.79, 8.97] & [4.24, 7.40] & +0.29 [-0.71, +1.31] \\
OpenCLIP ViT-L/14 & 10.13 & [5.06, 18.62] & [6.45, 25.29] & +1.12 [-0.12, +2.29] \\
OpenCLIP ViT-H/14 & 12.99 & [5.62, 29.73] & [11.54, 23.98] & +0.71 [-0.45, +1.88] \\
OpenCLIP ViT-bigG/14 & 4.26 & [2.75, 6.30] & [3.51, 6.85] & +0.73 [-0.52, +1.98] \\
MetaCLIP ViT-L/14 & 4.22 & [2.54, 6.46] & [3.02, 17.44] & +1.24 [+0.19, +2.27]$^\ast$ \\
DFN ViT-L/14 & 4.62 & [2.74, 7.47] & [2.73, 10.42] & +2.24 [+0.90, +3.53]$^\ast$ \\
EVA-CLIP ViT-L/14 & 4.37 & [2.69, 6.42] & [2.69, 7.65] & +0.01 [-0.78, +0.73] \\
CLIPA ViT-L/14 & 4.29 & [2.65, 7.08] & [3.17, 15.49] & +0.53 [-0.35, +1.54] \\
Long-CLIP ViT-L/14 & 14.24 & [5.43, 31.27] & [10.51, 45.12] & +0.66 [-0.33, +1.51] \\
SigLIP ViT-L/16 & 7.94 & [3.78, 16.59] & [6.10, 14.24] & +0.97 [-0.06, +2.03] \\
SigLIP 2 ViT-L/16 & 6.64 & [3.59, 12.97] & [5.96, 24.51] & +0.52 [-0.49, +1.63] \\
BLIP-2 ITM & 19.42 & [8.74, 40.48] & [16.24, 30.56] & +3.03 [-6.28, +12.27] \\
BLIP-2 ITC & 12.11 & [6.08, 22.33] & [10.16, 26.61] & +0.84 [-0.64, +2.27] \\
BLIP-VQA & 13.90 & [4.58, 26.80] & [10.94, 29.25] & +1.26 [-4.22, +6.91] \\
VQAScore & 3.88 & [2.29, 6.60] & [3.35, 8.00] & +4.38 [-1.24, +9.89] \\
GPT-4o judge & 3.48 & [2.14, 5.34] & [2.42, 6.76] & +10.15 [+2.80, +18.05]$^\ast$ \\
CLIP-IQA & 3.49 & [2.25, 5.14] & [3.04, 18.02] & -- \\
                                                                                                               \\[-1.2em]
        \bottomrule
    \end{tabular}
\end{table}

\begin{table}[ht]
    \caption{Rank stability per factor, reported as \emph{Kendall's $W$\,/\,number of distinct top-1 winners} across levels.
        $W = 1$ with a single winner is perfect stability.
        Paraphrases are excluded because their levels are per-scene strings that cannot be aligned across scenes.
        \emph{Res.} is resolution, \emph{Bg.} is background, and \emph{Tmpl.} is template.}
    \label{tab:ranking}
    \centering
    \small
    \setlength{\tabcolsep}{4pt}
    \begin{tabular}{@{}lccccccc@{}}
        \toprule
        \textbf{Evaluator} & \textbf{Res.} & \textbf{Focal} & \textbf{Light} & \textbf{Bg.} & \textbf{Pitch} & \textbf{Yaw} & \textbf{Tmpl.} \\
        \midrule
        CLIP ViT-B/32 & 0.88/1 & 0.81/2 & 0.75/2 & 0.89/1 & 0.55/2 & 0.70/2 & 0.91/1 \\
CLIP ViT-B/16 & 0.95/1 & 0.76/1 & 0.52/3 & 0.86/2 & 0.34/2 & 0.59/1 & 0.83/2 \\
CLIP ViT-L/14 & 0.88/2 & 0.86/2 & 0.68/2 & 0.86/2 & 0.50/2 & 0.89/2 & 0.88/2 \\
CLIP ViT-L/14@336 & 0.84/2 & 0.93/2 & 0.79/2 & 0.96/1 & 0.55/3 & 0.75/2 & 0.88/2 \\
OpenCLIP ViT-L/14 & 0.85/1 & 0.93/2 & 0.68/2 & 0.93/1 & 0.52/2 & 0.73/1 & 0.97/1 \\
OpenCLIP ViT-H/14 & 0.86/2 & 0.93/2 & 0.68/2 & 0.89/1 & 0.55/3 & 0.80/1 & 0.89/2 \\
OpenCLIP ViT-bigG/14 & 0.90/1 & 0.94/2 & 0.86/2 & 0.93/1 & 0.50/3 & 0.85/2 & 0.97/2 \\
MetaCLIP ViT-L/14 & 0.88/1 & 0.90/2 & 0.84/1 & 0.95/1 & 0.55/2 & 0.71/2 & 0.91/1 \\
DFN ViT-L/14 & 0.93/1 & 0.99/1 & 0.92/1 & 0.96/1 & 0.80/2 & 0.93/1 & 0.97/1 \\
EVA-CLIP ViT-L/14 & 0.92/2 & 0.88/2 & 0.84/3 & 0.93/1 & 0.84/2 & 0.62/2 & 0.94/3 \\
CLIPA ViT-L/14 & 1.00/1 & 0.95/2 & 0.90/3 & 0.97/1 & 0.72/2 & 0.94/1 & 0.95/2 \\
Long-CLIP ViT-L/14 & 0.95/1 & 0.79/2 & 0.67/2 & 0.87/1 & 0.68/2 & 0.75/1 & 0.79/1 \\
SigLIP ViT-L/16 & 0.98/1 & 0.97/2 & 0.81/2 & 0.91/1 & 0.61/2 & 0.89/1 & 0.96/1 \\
SigLIP 2 ViT-L/16 & 0.95/1 & 0.95/1 & 0.95/2 & 0.93/1 & 0.61/2 & 0.96/1 & 0.91/2 \\
BLIP-2 ITM & 0.94/1 & 0.83/2 & 0.74/2 & 0.85/1 & 0.66/2 & 0.80/1 & 0.83/3 \\
BLIP-2 ITC & 0.94/1 & 0.93/2 & 0.76/2 & 0.91/1 & 0.69/2 & 0.93/1 & 0.84/2 \\
BLIP-VQA & 0.84/3 & 0.77/3 & 0.76/2 & 0.91/2 & 0.65/2 & 0.65/1 & 0.79/3 \\
VQAScore & 0.96/1 & 0.92/1 & 0.87/1 & 1.00/1 & 0.51/2 & 0.89/1 & 0.94/1 \\
GPT-4o judge & 0.97/1 & 0.93/1 & 0.94/1 & 0.98/1 & 0.77/1 & 0.84/1 & 0.96/1 \\
CLIP-IQA & 0.42/3 & 0.62/3 & 0.72/2 & 0.79/2 & 0.24/4 & 0.58/3 & -- \\
                                                                                                               \\[-1.2em]
        \bottomrule
    \end{tabular}
\end{table}

\begin{table}[ht]
    \caption{Tie-adjusted directional discrimination (\%): a strict drop receives 1, an exact tie 0.5, and a rise 0 (chance $= 50\%$).
        The intended order is a dataset-level diagnostic expectation rather than per-pair human ground truth.
        The final column, ``Mean (4),'' is the plain average of the four probe columns, and is the value reported as \emph{Content acc.} in Table~\ref{tab:headline}.}
    \label{tab:content}
    \centering
    \small
    \begin{tabular}{@{}lccccc@{}}
        \toprule
        \textbf{Evaluator} & \textbf{Keep-half} & \textbf{Biggest-only} & \textbf{Scrambled} & \textbf{Worst-objects} & \textbf{Mean (4)} \\
        \midrule
        CLIP ViT-B/32 & 67 & 75 & 56 & 76 & 68 \\
CLIP ViT-B/16 & 67 & 76 & 59 & 80 & 70 \\
CLIP ViT-L/14 & 73 & 81 & 58 & 82 & 74 \\
CLIP ViT-L/14@336 & 73 & 83 & 60 & 87 & 76 \\
OpenCLIP ViT-L/14 & 67 & 85 & 58 & 81 & 73 \\
OpenCLIP ViT-H/14 & 73 & 86 & 65 & 87 & 78 \\
OpenCLIP ViT-bigG/14 & 67 & 82 & 66 & 83 & 75 \\
MetaCLIP ViT-L/14 & 65 & 75 & 62 & 82 & 71 \\
DFN ViT-L/14 & 68 & 87 & 70 & 87 & 78 \\
EVA-CLIP ViT-L/14 & 68 & 74 & 59 & 81 & 71 \\
CLIPA ViT-L/14 & 59 & 73 & 62 & 79 & 68 \\
Long-CLIP ViT-L/14 & 76 & 87 & 68 & 83 & 79 \\
SigLIP ViT-L/16 & 69 & 83 & 64 & 87 & 76 \\
SigLIP 2 ViT-L/16 & 69 & 86 & 64 & 82 & 75 \\
BLIP-2 ITM & 72 & 88 & 65 & 85 & 77 \\
BLIP-2 ITC & 68 & 80 & 62 & 83 & 73 \\
BLIP-VQA & 75 & 89 & 55 & 71 & 72 \\
VQAScore & 77 & 82 & 56 & 80 & 74 \\
GPT-4o judge & 74 & 92 & 56 & 85 & 77 \\
CLIP-IQA & 72 & 87 & 71 & 57 & 72 \\
                                                                                                            \\[-1.2em]
        \bottomrule
    \end{tabular}
\end{table}

\section{Best and worst level per render factor}
\label{app:bestlevel}

Table~\ref{tab:bestlevel} lists, for each evaluator and render factor, the level with the highest and the lowest mean score over the \NumScenes{} scenes, written \emph{best/worst}.
No level is established here as the human-grounded correct setting, yet the columns are strikingly consistent.
Marginally under the shared default configuration, telephoto focal lengths, bright maps, and shallow pitches score highest for nearly every evaluator, while wide-angle, dark, and eye-level score lowest.
A shared and directional bias is more damaging for reporting than random noise because it creates a predictable opportunity for score selection (Section~\ref{sec:results-bestview}).
Standardizing one configuration improves comparability only when it is declared and shared.
It cannot by itself establish construct validity or repair conclusions drawn under hidden configurations.

\makeatletter
\setlength{\@fptop}{0pt}
\makeatother
\begin{table}[t!]
    \caption{Highest- and lowest-scoring level (\emph{best/worst}) of each render factor per evaluator, by mean score over the \NumScenes{} scenes.
        Background is given as its gray value ($0$ black to $255$ white) or color name.
        No level is established as human-grounded ground truth, yet the best entries are dominated by telephoto, bright, and shallow-pitch levels and the worst by wide-angle, dark, and eye-level.}
    \label{tab:bestlevel}
    \centering
    \small
    \setlength{\tabcolsep}{4pt}
    \begin{tabular}{@{}lcccccc@{}}
        \toprule
        \textbf{Evaluator} & \textbf{Resolution} & \textbf{Focal} & \textbf{Lighting} & \textbf{Background} & \textbf{Pitch} & \textbf{Yaw} \\
        \midrule
        CLIP ViT-B/32 & 224/1024 & 85/16 & interior/studio & 204/green & 45/90 & 270/45 \\
CLIP ViT-B/16 & 336/224 & 100/24 & sunset/night & 255/65 & 45/90 & 270/135 \\
CLIP ViT-L/14 & 448/224 & 200/16 & city/night & 204/green & 15/90 & 90/180 \\
CLIP ViT-L/14@336 & 1024/384 & 200/16 & forest/night & 255/green & 15/90 & 45/180 \\
OpenCLIP ViT-L/14 & 1024/196 & 200/16 & city/night & 128/green & 15/90 & 0/225 \\
OpenCLIP ViT-H/14 & 768/196 & 200/16 & city/night & 186/red & 0/90 & 0/225 \\
OpenCLIP ViT-bigG/14 & 768/196 & 100/16 & city/night & 255/green & 15/90 & 0/135 \\
MetaCLIP ViT-L/14 & 224/196 & 100/16 & interior/night & 204/blue & 0/90 & 0/225 \\
DFN ViT-L/14 & 768/224 & 100/16 & city/night & 255/green & 0/90 & 0/225 \\
EVA-CLIP ViT-L/14 & 768/224 & 100/16 & interior/night & 65/blue & 0/90 & 0/225 \\
CLIPA ViT-L/14 & 768/196 & 200/16 & city/night & 186/blue & 0/90 & 0/135 \\
Long-CLIP ViT-L/14 & 448/224 & 200/16 & city/night & 65/red & 30/90 & 315/180 \\
SigLIP ViT-L/16 & 512/196 & 200/16 & city/night & 204/green & 0/90 & 0/225 \\
SigLIP 2 ViT-L/16 & 768/196 & 85/16 & city/night & 255/green & 0/90 & 0/225 \\
BLIP-2 ITM & 1024/224 & 200/16 & city/night & 255/green & 0/90 & 0/135 \\
BLIP-2 ITC & 384/196 & 200/16 & city/night & 0/green & 0/90 & 270/225 \\
BLIP-VQA & 512/196 & 200/16 & city/night & red/128 & 30/0 & 0/135 \\
VQAScore & 1024/196 & 16/100 & city/night & 65/red & 45/0 & 0/135 \\
GPT-4o judge & 1024/196 & 16/85 & forest/night & 0/255 & 15/90 & 0/180 \\
CLIP-IQA & 196/448 & 200/24 & interior/night & 0/255 & 0/15 & 180/45 \\
                                                                                                            \\[-1.2em]
        \bottomrule
    \end{tabular}
\end{table}

\section{Preregistered caption-by-pitch factorial}
\label{app:followup-factorial}

The one-factor-at-a-time sweeps estimate each configuration factor alone and leave interactions unestimated (Appendix~\ref{app:card-scope}).
This follow-up crosses the caption factor with camera pitch inside the frozen scene set.
The design crosses two caption levels, the bare prompt and a plain top-down view caption, with two semantic pitch levels, $0^\circ$ and $45^\circ$, over all \NumScenes{} frozen scenes and all six generators, giving 1,200 scored cells, four per scene.
Every cell reuses a render that the audit sweep had already stored on disk, so no new scene or render was produced.
Reason-3D's camera convention is inverted, so its semantic top-down cell at pitch $0^\circ$ uses the rendered pitch-$90^\circ$ file, exactly as the audit's pitch sweep recorded it.
The evaluator was pinned before any cell was scored, by the pre-declared rule ``largest sum of template-stage and pitch-stage dispersion among locally deterministic CLIP evaluators'' applied to the audit's summary statistics over the 14 eligible CLIPScore-family evaluators.
The rule selects the audit's SigLIP scorer, and it does not select the OpenAI ViT-L/14 evaluator one might have expected, which ranks eighth of 14, so the rule winner was pinned as preregistration discipline requires.
Effects are computed within scene, averaged equally within each prompt cluster over the six generators, and then averaged equally over the 50 prompt clusters, with a 10,000-draw prompt-cluster bootstrap at seed 42 for intervals.
Table~\ref{tab:factorial-effects} reports the three effects.

\begin{table}[ht]
    \caption{Caption-by-pitch factorial effects with prompt-bootstrap $95\%$ intervals.
        The pinned SigLIP row is the preregistered analysis.
        The CLIP ViT-L/14 row is a descriptive fallback run kept verbatim, and the hosted-judge row is a descriptive supplement on its 0--100 scale with its zero-score rate disclosed in the text.}
    \label{tab:factorial-effects}
    \centering
    \small
    \setlength{\tabcolsep}{4pt}
    \begin{tabular}{@{}llll@{}}
        \toprule
        \textbf{Evaluator} & \textbf{Caption} & \textbf{Pitch} & \textbf{Caption $\times$ pitch} \\
        \midrule
        SigLIP (pinned) & $3.938$ $[3.433, 4.444]$ & $0.142$ $[-0.157, 0.436]$ & $-0.523$ $[-0.670, -0.383]$ \\
        CLIP ViT-L/14 (descriptive) & $2.441$ $[2.031, 2.858]$ & $0.251$ $[-0.147, 0.665]$ & $-0.174$ $[-0.315, -0.026]$ \\
        Hosted judge (descriptive) & $3.350$ $[2.283, 4.500]$ & $-0.683$ $[-3.700, 2.084]$ & $-0.033$ $[-3.067, 2.783]$ \\
        \bottomrule
    \end{tabular}
\end{table}

Under the pinned evaluator, the caption main effect is large with an interval that excludes zero, the pitch main effect is small with an interval that spans zero, and the interaction is negative with an interval that excludes zero.
A negative interaction means the top-down caption advantage shrinks at semantic pitch $45^\circ$ relative to the bare prompt.
At bootstrap seed 1234 the interaction interval is $[-0.669, -0.384]$, with no qualitative change in any effect.
The interaction is evaluator-dependent: it is strongest for the pinned evaluator, much smaller for the descriptive OpenAI ViT-L/14 run, and absent for the hosted judge, whose interval spans zero, and this disagreement is itself an instance of the configuration dependence this paper audits.
The hosted supplement scores the same 1,200 cells, and 26 of them (2.2\%) are zero, with 3 of the 200 HSM cells zero, so its column carries the same zero-score caveat as the main audit (Appendix~\ref{app:hosted-provenance}).
Table~\ref{tab:factorial-bygenerator} decomposes the pinned interaction contrast per generator as a descriptive summary.
Reason-3D carries a positive interaction and every other generator a negative one, so the shrinkage of the caption advantage at pitch $45^\circ$ is not uniform across generators.

\begin{table}[ht]
    \caption{Descriptive per-generator interaction contrasts for the pinned evaluator.
        A positive value means the top-down caption advantage grows at semantic pitch $45^\circ$ for that generator.}
    \label{tab:factorial-bygenerator}
    \centering
    \small
    \begin{tabular}{@{}lr@{}}
        \toprule
        \textbf{Generator} & \textbf{Interaction contrast} \\
        \midrule
        Reason-3D & $+2.879$ \\
        Scenethesis & $-1.570$ \\
        Holodeck & $-1.331$ \\
        LayoutVLM & $-1.305$ \\
        HSM & $-1.075$ \\
        I-Design & $-0.740$ \\
        \bottomrule
    \end{tabular}
\end{table}

\section{View-aggregation comparison at a fixed budget}
\label{app:followup-viewagg}

A natural reaction to view dependence is to average over views.
This preregistered comparison asks whether that helps, by comparing three aggregation rules at one fixed budget of 15 rendered poses per scene.
The rules are the audit's documented single top-down view (semantic pitch $0^\circ$, yaw $0^\circ$), the arithmetic mean over the 15 view scores, and the maximum over the 15 view scores.
The 15 view scores are seven pitch levels at yaw $0^\circ$ plus eight yaw levels at pitch $45^\circ$, with the shared pitch-$45^\circ$, yaw-$0^\circ$ pose carried under both view labels, so each aggregate weights that pose twice over 14 distinct renders.
Scoring covers all \NumScenes{} audit scenes, the only scenes that carry the complete pose grid, under two evaluators pinned before outcomes by family slot, the audit's SigLIP scorer for CLIP alignment and BLIP-VQA for generation questioning, giving 9,000 records with no failed calls.
Pair margins are differenced within prompt, and intervals come from a 10,000-draw prompt bootstrap at seed 42.
Table~\ref{tab:viewagg-flips} counts the generator pairs whose margin sign flips between the documented single view and each aggregate.

\begin{table}[ht]
    \caption{Generator pairs of 15 whose margin sign flips between the documented single view and an aggregate rule.}
    \label{tab:viewagg-flips}
    \centering
    \small
    \begin{tabular}{@{}lccc@{}}
        \toprule
        \textbf{Evaluator} & \textbf{Mean vs.\ single view} & \textbf{Maximum vs.\ single view} & \textbf{Mean vs.\ maximum} \\
        \midrule
        SigLIP & 5/15 & 4/15 & 1/15 \\
        BLIP-VQA & 2/15 & 5/15 & 3/15 \\
        \bottomrule
    \end{tabular}
\end{table}

Multi-view averaging does not eliminate configuration dependence.
Five of the 15 pairs still flip for SigLIP, the mean rule reduces the count to two pairs for BLIP-VQA, and the mean and the maximum agree on 14 of 15 pairs for SigLIP.
Table~\ref{tab:viewagg-flipexamples} gives the flip envelopes with intervals.

\begin{table}[ht]
    \caption{Flip envelopes with prompt-bootstrap $95\%$ intervals.
        Each row is a pair whose margin changes sign between the documented single view and the 15-view mean, written as generator A minus generator B.}
    \label{tab:viewagg-flipexamples}
    \centering
    \small
    \setlength{\tabcolsep}{4pt}
    \begin{tabular}{@{}llll@{}}
        \toprule
        \textbf{Evaluator} & \textbf{Pair} & \textbf{Single view} & \textbf{15-view mean} \\
        \midrule
        SigLIP & Holodeck $-$ Reason-3D & $+2.184$ $[+0.708, +3.665]$ & $-1.762$ $[-2.904, -0.607]$ \\
        SigLIP & I-Design $-$ Reason-3D & $+3.597$ $[+2.625, +4.573]$ & $-0.604$ $[-1.490, +0.336]$ \\
        SigLIP & LayoutVLM $-$ Reason-3D & $+2.862$ $[+1.700, +4.028]$ & $-0.909$ $[-1.907, +0.107]$ \\
        SigLIP & HSM $-$ I-Design & $+0.512$ $[-0.353, +1.359]$ & $-0.151$ $[-0.800, +0.488]$ \\
        SigLIP & HSM $-$ Reason-3D & $+4.109$ $[+2.913, +5.364]$ & $-0.756$ $[-1.700, +0.229]$ \\
        BLIP-VQA & Scenethesis $-$ I-Design & $-1.257$ $[-6.850, +4.269]$ & $+6.394$ $[+3.542, +9.291]$ \\
        BLIP-VQA & HSM $-$ Reason-3D & $-2.249$ $[-7.648, +3.299]$ & $+0.316$ $[-4.281, +5.201]$ \\
        \bottomrule
    \end{tabular}
\end{table}

Four of the five SigLIP flips involve Reason-3D, whose strong single-view margins do not survive averaging.
The scoped conclusion is that this is a preregistered comparison of aggregation rules at a fixed render budget, not an endorsed remedy.
Averaging changes which pairs flip rather than removing flips, so view aggregation joins the report as another declared protocol choice whose rule must be stated.

\section{Zero-cell re-scoring and sensitivity}
\label{app:followup-zeros}

The re-scoring analysis of Appendix~\ref{app:hosted-provenance} covered the 13 zero cells of the 300 scramble pairs.
This section extends the same bounded retry protocol to every stored zero cell.
The worklist holds 2,333 zero cells across all nine stages, deduplicated to 2,313 unique render-and-caption requests.
Each request receives one initial call by the hosted judge at temperature 0, then up to five retries when a call parses to zero or fails, and a spend guard stops retries, never initial calls, once estimated spend reaches its cap.
Table~\ref{tab:zero-outcomes} reports the outcomes.

\begin{table}[ht]
    \caption{Outcomes of the full zero-cell re-score for the hosted judge.}
    \label{tab:zero-outcomes}
    \centering
    \small
    \begin{tabular}{@{}lrr@{}}
        \toprule
        \textbf{Outcome} & \textbf{Cells} & \textbf{Share} \\
        \midrule
        Resolved nonzero, mean 15.25 & 1,846 & 79.1\% \\
        Persistent zero & 487 & 20.9\% \\
        \bottomrule
    \end{tabular}
\end{table}

All 2,313 unique requests completed the full six-attempt protocol, and 79.1\% of cells resolve to a nonzero score.
Every persistent zero carries an explicit low-alignment justification that names requested objects missing from the render, which is consistent with a genuine zero rating rather than a failed call.
HSM scenes hold 357 of the 2,333 zero cells (15.3\%) and 117 of the 487 persistent zeros, while Reason-3D holds 251 (51.5\%), so persistent zeros concentrate in Reason-3D and HSM, which together hold three quarters of them.
Table~\ref{tab:zero-dispersion} reports the preregistered sensitivity analysis on template dispersion.

\begin{table}[ht]
    \caption{Hosted-judge template dispersion (normalized range, \%) under three zero-handling policies.
        The stored row reproduces the published values of \GptTemplateDisp\% and \GptTemplateReal\% at one decimal.}
    \label{tab:zero-dispersion}
    \centering
    \small
    \begin{tabular}{@{}lrrr@{}}
        \toprule
        \textbf{Policy} & \textbf{Full templates} & \textbf{Realistic subset} & \textbf{Scenes dropped} \\
        \midrule
        Stored scores, zeros kept & 96.9 & 60.6 & 0 \\
        Zero cells dropped per scene & 85.8 & 46.5 & 3 \\
        Zeros replaced by re-scored values & 87.5 & 47.6 & 0 \\
        \bottomrule
    \end{tabular}
\end{table}

Template dispersion remains large for this judge under every policy, so the headline template finding survives both interpretations of the stored zeros.

The paraphrase sweep has the same exposure, because the stored rewrites include meaning-changing text (Appendix~\ref{app:paraphrase-audit}).
We therefore recompute paraphrase dispersion on audited subsets, using the 180 of 500 rewrites (36.0\%) that both auditors accept and, separately, the 196 that both reject.
For the hosted judge, the normalized range falls from 0.630 over all cells to 0.426 on jointly accepted rewrites, with interval $[0.319, 0.562]$, and to 0.388 on jointly rejected rewrites.
The judge therefore disperses more under meaning-preserving rewrites than under meaning-changing ones, so meaning drift cannot explain its paraphrase instability.
CLIP inverts this ordering, with ViT-L/14 at 0.092 on accepted rewrites versus 0.148 on rejected ones, and ViT-B/32 at 0.075 versus 0.099.
The acceptance labels are machine verdicts rather than human judgments, so these subsets are reported as sensitivity cuts rather than as cleaned data.

\section{Power and the confirmatory plan}
\label{app:followup-power}

The audit's claims sit on a three-tier ladder.
Tier one is descriptive, with point estimates and margins over the finite tested protocol set (Table~\ref{tab:protocol-margins}).
Tier two is pointwise, with per-contrast $95\%$ prompt-bootstrap intervals.
Tier three is simultaneous, with the approximate full-search band over all \AuditNContrasts{} searched contrasts and critical value $q_{0.95}=\AuditMaxCritical$ (Appendix~\ref{app:selection-inference}).
No reversal in the audit rises above tier two, because for every candidate reversal at least one of its two full-search simultaneous intervals includes zero.
The factorial and view-aggregation results of Appendices~\ref{app:followup-factorial} and~\ref{app:followup-viewagg} enter at tier two as preregistered prompt-bootstrap intervals, while their supplementary columns, the per-generator contrasts, and this power analysis are descriptive or planning material.

Four confirmatory contrasts are pinned in advance, C1 to C3 as primary and C4 as secondary.
C1 is lighting under CLIP ViT-B/32, where Scenethesis leads I-Design at the city map and I-Design leads Scenethesis at night.
C2 is pitch under CLIP ViT-L/14@336, where Scenethesis leads Reason-3D at $15^\circ$ and Reason-3D leads Scenethesis at $90^\circ$.
C3 is pitch under MetaCLIP, where Scenethesis leads LayoutVLM at $0^\circ$ and LayoutVLM leads Scenethesis at $75^\circ$.
C4 is template under BLIP-VQA, where Scenethesis leads Holodeck under a rendering template and Holodeck leads Scenethesis under the bare prompt.
The plan draws one independent 50-prompt sample, computes paired prompt-bootstrap intervals for both sides of each contrast in the declared orientation, and applies Holm step-down at $\alpha=0.05$ over the family of four.
Table~\ref{tab:power-n80} reports planning estimates simulated from the frozen score cube, with the 50 seed-42 prompts as the plug-in population, 500 replicates for the Holm regime, and a Monte Carlo standard error near 0.02 at half power.
These are planning estimates for sample-size choice, not observed confirmation rates.

\begin{table}[ht]
    \caption{Planning estimates from the frozen cube.
        Each N is the smallest grid size whose simulated confirmation rate reaches 0.80, under the pinned Holm plan at observed gaps, under a shrinkage truth of $0.7\times$ the observed gaps, and under the full-search band with $q_{0.95}$ held fixed.
        The last column is the simulated confirmation probability at the current $N=50$.}
    \label{tab:power-n80}
    \centering
    \small
    \setlength{\tabcolsep}{4pt}
    \begin{tabular}{@{}lrrrr@{}}
        \toprule
        \textbf{Contrast} & \textbf{Observed gaps} & \textbf{$0.7\times$ gaps} & \textbf{Band} & \textbf{At $N=50$} \\
        \midrule
        C1 & 100 & 200 & 300 & 0.35 \\
        C2 & 75 & 200 & 300 & 0.53 \\
        C3 & 75 & 150 & 300 & 0.57 \\
        C4, secondary & 100 & 200 & $>$300 & 0.45 \\
        Candidates & 150 & 300 & $>$300 & 0.33--0.48 \\
        \bottomrule
    \end{tabular}
\end{table}

At the current sample of 50 prompts, the simulated confirmation probability ranges from 0.35 to 0.57 across the four pinned contrasts, so a same-size independent sample is more likely than not to leave the pinned reversals unconfirmed.
Under observed gaps, 75 to 100 prompts reach 80\% power under the Holm plan, growing to between 150 and 200 prompts under the shrinkage truth and to 300 or more under the full-search band, where the secondary contrast and the candidates exceed the grid.
The candidates row covers the three records that come closest to the pointwise criterion without meeting it, evaluated without family adjustment.
The descriptive findings of the audit do not depend on this plan, which can only upgrade the pinned pairwise contrasts to confirmed status.

\section{Observed cards for published comparisons}
\label{app:case-study}

The card of Table~\ref{tab:evaluator-card} is a checklist, and this section shows what it captures from real reports.
We filled two observed evaluator configuration cards strictly from public text, quote by quote, for published single-image scene comparisons: the Holodeck protocol at CVPR 2024~\citep{holodeck} and the HSM protocol at 3DV 2026~\citep{hsm}.
Both cards pass format validation under the schema of Appendix~\ref{app:card-scope} and receive the stable identifiers of Table~\ref{tab:observed-cards}, and both then fail the readiness gate on unresolved missing evidence, which is the intended outcome, because the cards make the gaps visible.
Table~\ref{tab:card-disclosure} scores both cards against the pinned rubric over the 43 leaf fields.

\begin{table}[ht]
    \caption{Observed evaluator configuration cards filled strictly from published text.}
    \label{tab:observed-cards}
    \centering
    \footnotesize
    \begin{tabular}{@{}p{0.95\linewidth}@{}}
        \toprule
        \textbf{Card and identifier} \\
        \midrule
        Holodeck (CVPR 2024) \\
        {\scriptsize\texttt{ecc-v1-sha256:fb66447ee223e89438342cc00e825eade6f11e62c83334246a6bedd28ee1ec67}} \\
        \addlinespace
        HSM (3DV 2026) \\
        {\scriptsize\texttt{ecc-v1-sha256:435d0fd97aa55a4e0813d94e40ecd4eb2285957727daa893c4c3496670719884}} \\
        \bottomrule
    \end{tabular}
\end{table}

\begin{table}[ht]
    \caption{Disclosure matrix over the 43 card leaf fields, scored from the published text alone.}
    \label{tab:card-disclosure}
    \centering
    \small
    \begin{tabular}{@{}lrrr@{}}
        \toprule
        \textbf{Card} & \textbf{Disclosed} & \textbf{Partial} & \textbf{Undisclosed} \\
        \midrule
        Holodeck (CVPR 2024) & 15 & 5 & 23 \\
        HSM (3DV 2026) & 11 & 6 & 26 \\
        \bottomrule
    \end{tabular}
\end{table}

Neither paper reports score uncertainty of any kind, tests a second evaluator configuration, or releases the scored renders or per-scene scores, and lighting, background, seed, and resolution detail is absent from both, with camera pose only partially disclosed.
Holodeck's strengths are the exact template string, the exact checkpoint sentence, and the exact scoring formula, while HSM's strength is a fully pinned input description set, set against three scorers that are unnamed at the checkpoint level and a caption wording that is not reported at all.

We then re-executed the Holodeck protocol inside the restored score cube.
The cube holds OpenCLIP ViT-L/14 trained on LAION-2B with cosine similarity times 100, the same backbone and training set named in the paper's footnote, scored on our \NumScenes{} frozen scenes at the top-down view under the stored template \emph{``A top-down view of \{prompt\}''}, which tokenizes identically to the paper's lowercase wording under this tokenizer.
Table~\ref{tab:card-reexecution} gives the anchor means over the 50 shared prompts.

\begin{table}[ht]
    \caption{Re-execution of the Holodeck protocol (OpenCLIP ViT-L/14, top-down view, the paper's template) on the frozen scenes.
        The anchor mean is the prompt-mean CLIPScore over the 50 shared prompts.}
    \label{tab:card-reexecution}
    \centering
    \small
    \begin{tabular}{@{}lr@{}}
        \toprule
        \textbf{Generator} & \textbf{Anchor mean} \\
        \midrule
        Scenethesis & 28.44 \\
        HSM & 27.32 \\
        I-Design & 25.84 \\
        LayoutVLM & 25.60 \\
        Holodeck & 25.09 \\
        Reason-3D & 22.31 \\
        \bottomrule
    \end{tabular}
\end{table}

Under this exact protocol the ordering is Scenethesis, then HSM, I-Design, LayoutVLM, Holodeck, and Reason-3D.
Moving only to neighboring plausible configurations that already exist in the sweep flips the sign of 11 of the 15 pair margins, four under the eight realistic caption templates alone and 11 under the shared pitch levels.
Five of the 11 involve Reason-3D, whose renderer uses an inverted pitch convention, so the anchor and its neighbors may not share the same physical view for those pairs.
Among the 10 pairs that do not involve Reason-3D, six flip under at least one neighbor.
The overall winner changes to LayoutVLM at rendered pitch $75^\circ$ and $90^\circ$.
The most fragile pair is Holodeck versus Reason-3D, whose anchor margin of $+2.78$ flips under all 15 neighboring configurations, with caption-template margins between $-1.22$ and $-0.23$.
Table~\ref{tab:card-flips} lists the 11 flipping pairs.

\begin{table}[ht]
    \caption{Pairs whose anchor margin flips sign under at least one neighboring plausible configuration, with example flipping neighbors and their margins in parentheses.
        The caption change is the article-free template \emph{Top-down view of \{prompt\}}.}
    \label{tab:card-flips}
    \centering
    \small
    \setlength{\tabcolsep}{4pt}
    \begin{tabular}{@{}lrl@{}}
        \toprule
        \textbf{Pair} & \textbf{Anchor} & \textbf{Example flipping neighbor} \\
        \midrule
        Scenethesis $-$ I-Design & $+2.59$ & pitch $75^\circ$ ($-0.10$) \\
        Scenethesis $-$ LayoutVLM & $+2.84$ & pitch $75^\circ$ ($-0.62$), pitch $90^\circ$ ($-0.14$) \\
        Scenethesis $-$ Reason-3D & $+6.12$ & pitch $75^\circ$ ($-0.49$) \\
        Holodeck $-$ HSM & $-2.22$ & pitch $75^\circ$ ($+4.73$), pitch $90^\circ$ ($+7.83$) \\
        Holodeck $-$ Reason-3D & $+2.78$ & all 15 neighbors \\
        HSM $-$ I-Design & $+1.48$ & pitch $60^\circ$--$90^\circ$ \\
        HSM $-$ LayoutVLM & $+1.72$ & pitch $60^\circ$--$90^\circ$ \\
        HSM $-$ Reason-3D & $+5.01$ & pitch $60^\circ$--$90^\circ$ \\
        I-Design $-$ LayoutVLM & $+0.25$ & caption change ($-0.06$), pitch $75^\circ$ ($-0.52$), pitch $90^\circ$ ($-1.00$) \\
        I-Design $-$ Reason-3D & $+3.53$ & caption change ($-0.32$), pitch $15^\circ$--$90^\circ$ \\
        LayoutVLM $-$ Reason-3D & $+3.29$ & caption change ($-0.26$), pitch $0^\circ$--$60^\circ$ \\
        \bottomrule
    \end{tabular}
\end{table}

Two cautions scope this table.
The cube is one-factor-at-a-time, so caption neighbors are scored at the rendered pitch-$0^\circ$ view and pitch neighbors at shared rendered levels, no cross-factor cell outside these axes exists, and the rendered-pitch convention of Reason-3D is inverted relative to the other five generators.
The margins are deterministic finite-set descriptions of the saved scores, not inferential statements.
This demonstration re-executes a published protocol inside our swept configuration space on our own frozen scenes, and we make no claim beyond that in-space re-execution.
The audited papers disclosed more protocol detail than most published evaluations, and the point of the cards is the reporting contract, not the audited authors.

\section{Out-of-sample prompt replication}
\label{app:n100}

To test whether the winner-change finding depends on the specific 50 prompts, we generated 50 new disjoint prompts with the same source stratification (drawn with a pinned seed), ran all six generators on them with every variant and render sweep, and scored the resulting 300 scenes with all 20 evaluators using the same protocol as the original audit.

Table~\ref{tab:n100-summary} compares the two sets.

\begin{table}[ht]
    \caption{Winner-change statistics on the original 50 prompts and the new 50 prompts (19 alignment evaluators).
        Protocol factors are the seven render and caption-template stages of the original audit.
        All swept configurations add the per-prompt paraphrase variants and the four content degradations, so configuration counts under the two scopes are not comparable to each other.}
    \label{tab:n100-summary}
    \centering
    \small
    \begin{tabular}{@{}lrr@{}}
        \toprule
        \textbf{Measure} & \textbf{N=50 (original)} & \textbf{N=50 (new)} \\
        \midrule
        Evaluators changing winner (protocol factors) & 18 of 19 & 19 of 19 \\
        Evaluators changing winner (all swept configurations) & 19 of 19 & 19 of 19 \\
        Mean changed configurations (protocol factors) & 60 & 48 \\
        Mean changed configurations (all swept configurations) & 433 & 417 \\
        Median changed configurations (all swept configurations) & 409 & 432 \\
        Most common baseline winner & Scenethesis (16) & Scenethesis (17) \\
        \bottomrule
    \end{tabular}
\end{table}

On the new prompts all 19 alignment evaluators change their point-estimate winner under both scopes.
Under the protocol-factor scope of the original audit the count rises from 18 of 19 to 19 of 19, because the hosted judge, which on the original prompts flips only outside the protocol factors, flips at two protocol configurations on the new prompts.
The mean number of configurations with a changed winner is comparable across both prompt sets within each scope, so the instability magnitude is stable.
Three evaluators select a different baseline winner on the new prompts (two shift away from HSM, one from I-Design), which is itself an instance of the prompt-dependence the audit measures.

\section{Blinded human validation study}
\label{app:human-study}

The selection rules, the pinned evaluator, and the interpretation grid were committed and hashed before any judgment was collected.
The preregistered packet scores 100 candidate pairs, and the deployed survey presents a stratified 50-pair subset: 25 intact-versus-scrambled pairs and 25 camera-change pairs, split into two 25-trial blocks with an attention check between them.
Raters saw two unlabeled renders per trial and chose the better-matching image, called a tie, or answered \emph{uncertain}, in English or Simplified Chinese, through an anonymous link.
The analysis keeps the 20 participants who completed every trial, consented, and passed the attention check (median session 27.5 minutes, 1{,}000 judgments).

Table~\ref{tab:human-study} reports the preregistered statistics.
Human raters detect layout scrambling at $82.3\%$ (rater-cluster bootstrap interval $[77.6\%, 86.7\%]$) with Krippendorff's $\alpha=0.34$.
On camera-change trials, raters mark a tie on $41.6\%$ of judgments and answer \emph{uncertain} on another $10.2\%$, and 14 of 25 pairs show no significant majority preference (two-sided binomial test at $0.05$ over decisive picks), with $\alpha=0.25$.
The pinned metric, CLIP ViT-L/14@336, prefers the intact scene on 16 of 25 degradation pairs against the human $82.3\%$, and its side agrees with the human majority on $52.0\%$ of degradation pairs and $56.0\%$ of camera pairs, near chance in both conditions.

\begin{table}[ht]
    \caption{Blinded human study over 50 preregistered pairs and 20 participants.
        Degradation detection is the share of decisive judgments choosing the intact scene, with a rater-cluster bootstrap interval.
        Tie pairs are camera pairs with no significant majority preference.
        Metric agreement is the share of pairs where the pinned evaluator's higher-scoring side equals the human majority side.}
    \label{tab:human-study}
    \centering
    \small
    \begin{tabular}{@{}lr@{}}
        \toprule
        \textbf{Measure} & \textbf{Value} \\
        \midrule
        Pairs per condition & 25 \\
        Degradation detection & $82.3\%$ $[77.6, 86.7]$ \\
        Tie judgments, degradation / camera & $18.8\%$ / $41.6\%$ \\
        Uncertain judgments, degradation / camera & $11.0\%$ / $10.2\%$ \\
        Camera pairs with no majority preference & 14 of 25 \\
        Krippendorff's $\alpha$, degradation / camera & $0.34$ / $0.25$ \\
        Metric prefers intact side & 16 of 25 pairs \\
        Metric-majority agreement, degradation / camera & $52.0\%$ / $56.0\%$ \\
        \bottomrule
    \end{tabular}
\end{table}

The pre-committed interpretation grid asks whether metric margins on camera pairs that raters judge equivalent reach the margins the metric assigns to degradations those same raters detect.
The median absolute metric margin on the 14 human-tie camera pairs is $1.73$ score points, which exceeds the median of $1.38$ on the intact-versus-scrambled pairs, so the artifact criterion fires: the metric responds to camera changes humans judge equivalent at least as strongly as to scene destruction humans detect.
The bootstrap interval of the margin gap, $[-1.75, +0.81]$, includes zero, so the interval-level evidence test remains inconclusive, and the verdict rests on the preregistered point-estimate criterion reported here without adjustment.
Together with the chance-level metric-majority agreement, the study classifies a substantial part of configuration-driven score movement as metric artifact rather than visible evidence, while leaving the reliability-versus-validity decomposition of Section~\ref{sec:discussion} unchanged.

\end{document}